%% file: main.tex
\documentclass[pdflatex,sn-aps,Numbered]{sn-jnl}

\usepackage{placeins}
\usepackage{natbib}
\setcitestyle{square, comma, numbers,sort&compress, super}
\usepackage{subfig}
\usepackage{multirow, multicol, tabularx, longtable}
\usepackage{nicefrac}
\usepackage{siunitx}
\usepackage{array,framed}
\usepackage{booktabs}
\usepackage{
  color,
  float,
  epsfig,
  wrapfig,
  graphics,
  graphicx
}
\usepackage{verbatim}
\usepackage{textcomp}
\usepackage{setspace}
\usepackage{bm}
\usepackage{latexsym,fancyhdr,url}
\usepackage{enumerate}
\usepackage[ruled]{algorithm2e}
\usepackage{algpseudocode}
\usepackage{xparse} 
\usepackage{xspace}
\usepackage{csvsimple}
\usepackage{algpseudocode}

\usepackage{amsmath,amssymb,amsfonts}%
\usepackage{amsthm}%
\usepackage{mathrsfs}%
\usepackage[title]{appendix}%
\usepackage{xcolor}%
\usepackage{manyfoot}%
\usepackage{booktabs}%
\usepackage{listings}%
\usepackage{lscape}

\RequirePackage{amsthm}
\theoremstyle{thmstyleone}%
\theoremstyle{thmstyletwo}%

\theoremstyle{thmstylethree}%

\begin{document}

\title[Article Title]{MAPPING: Debiasing Graph Neural Networks for Fair Node Classification with Limited Sensitive Information Leakage}


\author*[1]{\sur{Ying Song}}\email{yis121@pitt.edu}

\author[1]{\sur{Balaji Palanisamy}}\email{bpalan@pitt.edu}


\affil*{\orgdiv{Department of Informatics and Networked Systems}, \orgname{University of Pittsburgh}, \orgaddress{\street{135 North Bellefield Avenue}, \city{Pittsburgh}, \postcode{15213}, \state{PA}, \country{USA}}}




\abstract{\input{Abstract}
}




\maketitle

\input{Introduction}
\input{Preliminaries}
\input{Empirical}
\input{Framework}
\input{Experiments}

\input{Related}
\input{Conclusion}
\bmhead{Acknowledgements}

This research was supported in part by the University of Pittsburgh Center for Research Computing through the resources provided. The first author acknowledges the support from the SCI fellowship.

\section*{Declarations}

\begin{itemize}
\item Funding: This research was supported in part by the University of Pittsburgh Center for Research Computing through the resources provided.
\item Conflict of interest/Competing interests: The authors have no relevant financial or non-financial interests to disclose.
\item Ethics approval and consent to participate: Yes.
\item Consent for publication: Yes.
\item Data availability: Yes. 
\item Materials availability: Yes. 
\item Code availability: Currently no.
\item Author contribution: Writing - Reviewing and editing and Supervision: Balaji Palanisamy; The rest: Ying Song
\end{itemize}


\bigskip
\begin{flushleft}%
Editorial Policies for:

\bigskip\noindent
Springer journals and proceedings: \url{https://www.springer.com/gp/editorial-policies}

\bigskip\noindent
Nature Portfolio journals: \url{https://www.nature.com/nature-research/editorial-policies}

\bigskip\noindent
\textit{Scientific Reports}: \url{https://www.nature.com/srep/journal-policies/editorial-policies}

\bigskip\noindent
BMC journals: \url{https://www.biomedcentral.com/getpublished/editorial-policies}
\end{flushleft}

\appendix
\input{Appendix}

\bibliography{reference}

\end{document}

%% file: Abstract.tex
Despite remarkable success in diverse web-based applications, Graph Neural Networks (GNNs) inherit and further exacerbate historical discrimination and social stereotypes, which critically hinder their deployments in high-stake domains such as online clinical diagnosis, financial crediting, etc. However, existing research in fair graph learning typically favors pairwise constraints to achieve fairness but fails to cast off dimensional limitations and generalize them into multiple sensitive attributes. Besides, most studies focus on in-processing techniques to enforce and calibrate fairness, constructing a model-agnostic debiasing GNN framework at the pre-processing stage to prevent downstream misuses and improve training reliability is still largely under-explored. Furthermore, previous work tends to enhance either fairness or privacy individually but few probes into how fairness issues trigger privacy concerns and whether such concerns can be alleviated with fairness intervention. In this paper, we propose a novel model-agnostic debiasing framework named MAPPING (\underline{M}asking \underline{A}nd \underline{P}runing and Message-\underline{P}assing train\underline{ING}) for fair node classification, in which we adopt the distance covariance ($dCov$)-based fairness constraints to simultaneously reduce feature and topology biases under multiple sensitive memberships, and combine them with adversarial debiasing to confine the risks of sensitive attribute inference. Experiments on real-world datasets with different GNN variants demonstrate the effectiveness and flexibility of MAPPING. Our results show that MAPPING can achieve better trade-offs between utility and fairness, and mitigate privacy risks of sensitive information leakage. This work paves the way for a new direction in trustworthy GNNs by addressing fairness and privacy concerns simultaneously, rather than achieving fairness at the expense of privacy.

%% file: Introduction.tex
\section{Introduction}

Graph Neural Networks (GNNs) have shown superior performance in various web applications, including recommendation systems \cite{social_recom} and online advertisement \cite{web_adv}. Message-passing schemes (MP) \cite{wu2019comprehensive_gnn} empower GNNs by aggregating node information from local neighborhoods, thereby rendering the clearer boundary between similar and dissimilar nodes \cite{fairwalk} to facilitate downstream graph tasks. However, disparities among different demographic groups can be perpetuated and amplified, causing severe social consequences in high-stake scenarios \cite{fairconsequences}. For instance, in clinical diagnosis \cite{gender_bias}, men are more extensively treated than women with the same severity of symptoms in a plethora of diseases, and older men aged 50 years or above receive extra healthcare or life-saving interventions than older women. With more GNNs adopted in medical analysis, gender discrimination may further deteriorate and directly cause misdiagnosis for women or even life endangerment.


Unfortunately, effective bias mitigation on non-i.i.d graphs is still largely under-explored, which particularly faces the following two problems. First, they tend to employ in-processing methods \cite{dai2021sayno_fair,nifty,learn_fairGNN,fmp_topology_bias}, which usually introduce complex fairness constraints, leading to high computational costs. Only a few models jointly alleviate feature and topology biases at the pre-processing stage and then feed debiased data into any GNN variants. This model-agnostic process is less likely to propagate biases under MP and is more flexible to deploy in real practice. For instance, FairDrop \cite{fairdrop} pre-modifies graph topologies to minimize distances among sensitive subgroups. However, it ignores the significant roles of node features to encode biases. Kamiran et al. \cite{pre_process_class} and Wang et al. \cite{counterfactual_pre_process} pre-debias features by ruffling, reweighting, or counterfactual perturbation, whereas their methods cannot be trivially applied to GNNs since unbiased features with biased topologies can result in biased distributions among different groups \cite{dong_fairgraph_survey}. Second, although recent studies \cite{dong2022edits} address the aforementioned gap, they introduce pairwise constraints, such as covariance ($Cov$), mutual information ($MI$), and Wasserstein distance ($Was$) \cite{optimal_transport}, to promote fairness. However, these methods are computationally inefficient in high dimensions and cannot be easily extended into multiple sensitive attributes. Besides, $Cov$ cannot reveal mutual independence between target variables and sensitive attributes; $MI$ cannot break dimensional limitations and be intractable to compute, and some popular estimators, e.g., MINE \cite{MINE} are proved to be heavily biased \cite{biasMI}; and Was is sensitive to outliers \cite{robust_wass_dis}, which hinders its uses in heavy-tailed data samples. To address these challenges, we utilize distance covariance ($dCov$), a distribution-free, scale-invariant \cite{dcor_free}, and outlier-resistant \cite{dcor_outliers} metric, as the fairness constraint. Most importantly, $dCov$ enables computations in arbitrary dimensions and ensures independence. We combine it with adversarial training to develop a feature and topology debiasing framework for GNNs.

Sensitive attributes not only exacerbate biases but also raise significant privacy concerns. These two factors impede the development of trustworthy GNNs, which require non-discrimination across subgroups based on sensitive attributes and the prevention of sensitive information leakage \cite{trustworthy_gnns}. Prior work has elucidated that GNNs are vulnerable to diverse types of attacks, such as attribute inference attacks, linking stealing attacks, and membership inference attacks \cite{GNNprivacy20, link_steal, nodelevel_MIA,label_MIA}. Even though identifiable information is masked and such pre-processed datasets are released in public for specific purposes, e.g., research institutions publish pre-processed real-world datasets for researchers to use, malicious third parties can still combine masked features with prior knowledge to recover sensitive attributes. In practice, links are preferentially connected based on sensitive attributes \cite{dong_fairgraph_survey}, which indicates that topological structures can contribute to sensitive attribute inferences and in turn, specific links can be stolen once sensitive attributes are identified. Furthermore, nodes typically carry multiple sensitive attributes rather than a single one, but a simple combination of multiple sensitive attributes (i.e., largely aligned with quasi-identifiers in privacy) can uniquely identify individuals \cite{unique_identify}, which may allow inferring other nodes' multiple sensitive attributes due to topological structures. To make things worse, some data samples, e.g., users or customers, may unintentionally disclose their multiple sensitive attributes to the public. Attackers can exploit these open resources to infer sensitive information, which enables further attacks and amplifying group inequalities, resulting in immeasurable social impacts. Thus, it is crucial to mitigate these privacy risks arising from features and topologies at the pre-processing stage under multiple sensitive attribute cases. 

PPFR \cite{interaction_priv_fair} is the first work to explore such interactions in GNNs, demonstrating both theoretically and empirically that improving individual fairness comes at the expense of increased edge-level privacy risks. However, it is not designed for multiple sensitive attributes and relies on post-processing techniques to achieve individual fairness. Yet this work motivates us to explore how fairness issues evolved from multiple sensitive attributes in GNNs exacerbate privacy risks and whether improving fairness can simultaneously reduce these risks. To address the above problems, we first conduct preliminary experiments both on synthetic and real-world datasets to empirically prove fairness issues can aggravate privacy risks of multiple sensitive attribute inferences. Next, we propose MAPPING with $dCov$-based constraints and adversarial training to decorrelate sensitive information from features and topologies. We evaluate privacy risks via attribute inference attacks and the empirical results showcase that MAPPING successfully guarantees fairness while ameliorating multiple sensitive information leakage, rather than making a careful trade-off between fairness and privacy. To the best of our knowledge, this is the first work to highlight the inner relationship between group-level fairness and multiple attribute privacy in GNNs at the pre-processing stage, contributing to the advancement of trustworthy GNNs. Please note that we aim to investigate how unfair GNNs can contribute to privacy risks and promote fairness in GNNs with limited sensitive information leakage. Providing rigorous privacy guarantees such as differential privacy \cite{sayno_privacy_extension,graph_emb_attr_infer_attack4recommendation,LPGNN} is out of our scope. 

In summary, our main contributions are threefold:

\textit{\textbf{MAPPING}} We propose a novel debiasing framework called MAPPING for fair node classification, which confines multiple sensitive attribute inferences derived from pre-debiased features and topologies. Our empirical results demonstrate that MAPPING obtains better flexibility and generalization to any GNN variants. 

\textit{\textbf{Effectiveness and Efficiency}} We evaluate MAPPING on three real-world datasets and compare MAPPING with vanilla GNNs and state-of-the-art debiasing models. The experimental results confirm the effectiveness and efficiency of MAPPING.

\textit{\textbf{Alignments and Trade-offs}} We explore the inner relationships between fairness and privacy in GNNs in the context of multiple sensitive attributes and illustrate MAPPING can achieve better trade-offs between utility and fairness while mitigating privacy risks of multiple sensitive attribute inferences.

%% file: Preliminaries.tex
\section{Preliminaries}
 In this section, we present the notations and introduce preliminaries of GNNs, $dCov$, two fairness metrics $\Delta SP$ and $\Delta EO$, and attribute inference attacks to measure sensitive information leakage.
\subsection{Notations}
In our work, we focus on node classification tasks. Given an undirected attributed graph $\mathcal{G}=\left(\mathcal{V}, \mathcal{E}, \mathcal{X}\right)$, where $\mathcal{V}$ denotes a set of nodes, $\mathcal{E}$ denotes a set of edges and the node feature set $\mathcal{X}=\left(\mathcal{X}_N, S\right)$ concatenates non-sensitive features $\mathcal{X}_N\in \mathcal{R}^{n\times d}$ and a sensitive attribute $S$. The goal of node classification is to predict the ground-truth labels $\mathcal{Y}$ as $\hat{Y}$ after optimizing the objective function $f_{\theta}(\mathcal{Y}, \hat{Y})$. Beyond the above notations, $A \in \mathbb{R}^{n \times n}$ is the adjacency matrix, where $n=|\mathcal{V}|$, $A_{ij}=1$ if two nodes $\left(v_i, v_j\right) \in \mathcal{E}$, otherwise, $A_{ij}=0$. 
\subsection{Graph Neural Networks} 
GNNs utilize MP to aggregate information of each node $v\in\mathcal{V}$ from its local neighborhood $\mathcal{N}(v)$ and thereby update its representation $H_{v}^{l}$ at layer $l$, which can be expressed as: 
\begin{equation}
    \setlength{\abovedisplayskip}{3pt}
    \setlength{\belowdisplayskip}{3pt}
    H_{v}^{l}=UPD^{l}(H_{v}^{l-1},AGG^{l-1}(\{H_{u}^{l-1}:u\in\mathcal{N}(v)\}))
\end{equation}
where $H_{v}^0=X_{v}$, and $UPD$ and $AGG$ are arbitrary differentiable functions that distinguish the different GNN variants. For a $l$-th layer GNN, typically, $H_{v}^{l}$ is fed into i.e., a linear classifier with a softmax function to predict node $v$'s label.
\subsection{Distance Covariance}
$dCov$ reveals independence between two random variables $X \in \mathbb{R}^p$ and $Y \in \mathbb{R}^q$ that follow any distribution, where $p$ and $q$ are arbitrary dimensions. As defined in \cite{dcororiginal}, given a sample $(X,Y)=\{(X_k,Y_k):k=1,\dots,n\}$ from a joint distribution, the empirical $dCov$ - $\mathcal{V}_n^2(X,Y)$ and its corresponding distance correlation ($dCor$) - $\mathcal{R}_n^2(X,Y)$ are defined as: 
\begin{equation}
    \setlength{\abovedisplayskip}{3pt}
    \setlength{\belowdisplayskip}{3pt}
    \begin{aligned}
        & \mathcal{V}_n^2(X,Y)=\frac{1}{n^2}\sum_{k,l=1}^{n}A_{kl}B_{kl} \\
        & \mathcal{R}_n^2(X,Y)=\begin{cases}
\frac{\mathcal{V}_n^2(X,Y)}{\sqrt{\mathcal{V}_n^2(X)\mathcal{V}_n^2(Y)}}, & \mathcal{V}_n^2(X)\mathcal{V}_n^2(Y)> 0 \\
0, & \mathcal{V}_n^2(X)\mathcal{V}_n^2(Y)=0
\end{cases}
    \end{aligned}    
\end{equation}
where n is the sampling number. $A_{kl}=a_{kl}-\Bar{a}_{k\cdot}-\Bar{a}_{\cdot l}+\Bar{a}_{\cdot \cdot}$, wherein the Euclidean distance matrix $a_{kl}=|X_k - X_l|_{p}$, $\Bar{a}_{k\cdot}=\frac{1}{n}\sum_{l=1}^{n}a_{kl}$, $\Bar{a}_{\cdot l}=\frac{1}{n}\sum_{k=1}^{n}a_{kl}$, and $\Bar{a}_{\cdot \cdot}=\frac{1}{n^2}\sum_{k,l=1}^{n}a_{kl}$ accordingly. $B_{kl}$ is defined similarly. The squared distance variance $\mathcal{V}_{n}^2(X)=\mathcal{V}_{n}^2(X,X)=\frac{1}{n^2}\sum_{k,l=1}^{n}A_{kl}^2$ and $\mathcal{V}_{n}^2(Y)$ is defined similarly. $\mathcal{V}_n^2(X,Y)\ge0$ and $0\le \mathcal{R}_n^2(X,Y)\le 1$. And $\mathcal{V}_n^2(X,Y)=\mathcal{R}_n^2(X,Y)=0$ iff $X$ and $Y$ are independent. For more details and corresponding proofs, please refer to \cite{dcororiginal}. 

\subsubsection{Links to Other Key Fairness/Privacy Constraints} The classic work \cite{dcororiginal} proves when two random variables $X$ and $Y$ jointly follow a bivariate normal distribution, $dCor$ is a deterministic function of Pearson correlation coefficient, which is the scaled form of $Cov$. \cite{fairalternative} shows $dCov$ is a tighter lower bound to MI. Under specific conditions, there is an asymptotic equivalence between MI and $dCov$. We refer the interested readers for more details in \cite{dcororiginal,fairalternative}.

\subsection{Fairness Metrics}
Given a binary label $\mathcal{Y} \in \left\{0,1\right\}$ and its predicted label $\hat{Y}$,
and a binary sensitive attribute $S \in \left\{0,1\right\}$, statistical parity (SP) \cite{fairthroughaware} and equal opportunity (EO) \cite{eo} can be defined as follows: 

\textit{\textbf{SP}} SP requires that $\hat{Y}$ and $S$ are independent, written as $P(\hat{Y} | S=0)=P(\hat{Y} | S=1)$, which indicates that the positive predictions between two subgroups should be equal.

\textit{\textbf{EO}} EO adds extra requirements for $\mathcal{Y}$, which requires the true positive rate between two subgroups to be equal, mathematically, $P(\hat{Y} | \mathcal{Y}=1, S=0)=P(\hat{Y} | \mathcal{Y}=1, S=1)$.

Following \cite{louizos2017variational_fair}, we use the differences of SP and EO between two subgroups as fairness measures, expressed as: 
\begin{equation}
    \setlength{\abovedisplayskip}{3pt}
    \setlength{\belowdisplayskip}{3pt}
    \begin{aligned}
        & \Delta SP = |P(\hat{Y} | S=0)-P(\hat{Y} | S=1)| \\
        & \Delta EO=|P(\hat{Y} | \mathcal{Y}=1, S=0)-P(\hat{Y} | \mathcal{Y}=1, S=1)|
    \end{aligned}
\end{equation}

\subsection{Attribute Inference Attacks}
Considering the alignment between fairness and privacy for multiple sensitive attributes, we naturally utilize attribute inference attacks to measure multiple sensitive information leakage. We assume adversaries can access pre-debiased features $\hat{X}$, topologies $\hat{A}$, and labels $\mathcal{Y}$, and gain partial multiple sensitive attributes $S_p$ through legal or illegal channels as prior knowledge. They integrate these sources to infer the target sensitive attributes $\hat{S}$. We assume they cannot tamper with internal parameters or architectures under the black-box setting. The attacker's goal is to train a supervised attack classifier $f_{\theta_{att}}(\hat{X},\hat{A},\mathcal{Y})=\hat{S}$ with any GNN variant to infer $\hat{S}$. This attack assumption is practical in real-world scenarios. For instance, business companies may provide model access via APIs, where adversaries are blocked from querying models since they cannot pass authentication or further be identified by detectors, but they can still download graph data from corresponding business-sponsored competitions in public platforms, e.g. Kaggle. Additionally, strict legal privacy and compliance policies in research or business institutions only allow partial employees to deal with sensitive data and then transfer pre-process information to other departments. While attackers cannot impersonate formal employees or access strongly confidential databases, they can steal sensitive information during routine communications. 

%% file: Empirical.tex
\section{Empirical Analysis}
\label{analysis}

In this section, we first utilize $dCor$ to investigate biases arising from node features and graph topology and can be amplified during MP. We empirically demonstrate that biased node features and/or topological structures can be fully exploited by malicious third parties to launch attribute inference attacks, which in turn can perpetuate and amplify existing social discrimination and stereotypes. We use synthetic datasets with multiple sensitive attributes to conduct these experiments. As suggested by Bose et al. \cite{compositional_fair}, we do not add any activation function to avoid nonlinear effects. We note that prior work \cite{dai2021sayno_fair} has demonstrated that topologies and MP can both exacerbate biases hidden behind node features. For instance, FairVGNN \cite{fairview} has illustrated that even masking sensitive attributes, sensitive correlations still exist after feature propagation. However, they are all evaluated on pairwise metrics of a single sensitive attribute. To evaluate the similar phenomena, we leverage a $2$-layer GNN to aggregate and update information twice. 
\subsection{Data Synthesis} 
\begin{figure}[t]
    \centering
    \subfloat[Biased non-sensitive features and graph topology (Major)]
    {
        \includegraphics[width=0.45\textwidth]{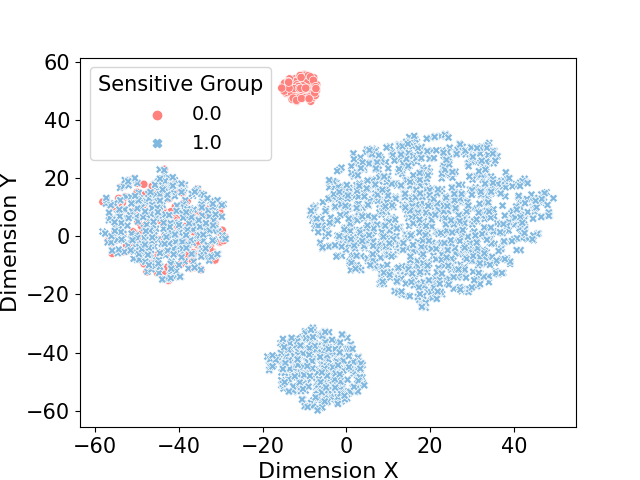}
        \includegraphics[width=0.45\textwidth]{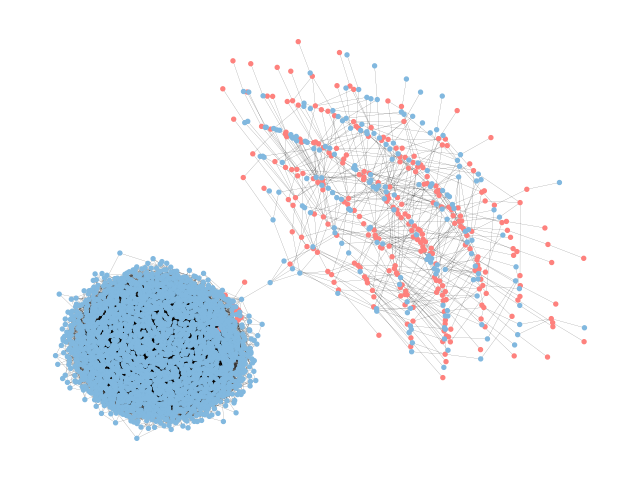}
        \label{Biased features and topology}}
    \hfill
    \subfloat[Unbiased non-sensitive features and graph topology (Major)]
    { 
    \includegraphics[width=0.45\textwidth]{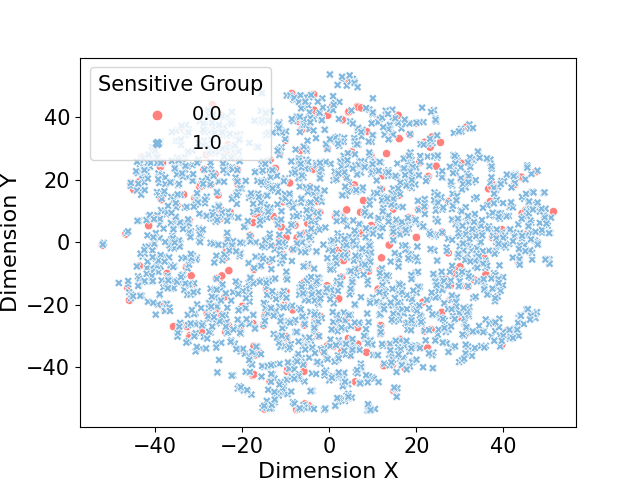}
    \includegraphics[width=0.45\textwidth]{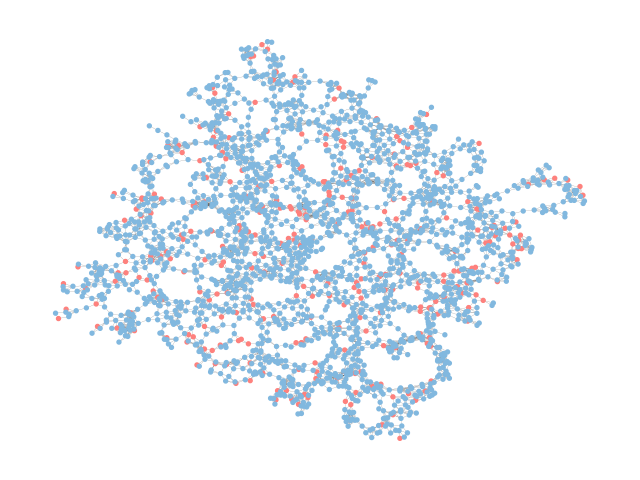}
        \label{Unbiased features and topology}}
    \caption{Distributions of Biased and Unbiased Graph Data Based on the Major Sensitive Attribute. The major sensitive attribute is binary, where group 0 represents the minority while group 1 denotes the majority.}
    \label{data_synthesis}
\end{figure}

First, we craft the main sensitive attribute $S_m$ and the minor $S_n$. E.g., to investigate racism in risk assessments \cite{recidivism}, `race' is the key focus while `age' follows behind. Please note that we only consider binary sensitive attributes for simplicity. We categorize the synthetic data into minority and majority groups based on the value of each sensitive attribute and utilize the distribution difference to reveal biases, in line with existing fairness studies \cite{fairthroughaware, dong2022edits}. An overview of distributions of the synthetic data (divided by the major sensitive attribute) is shown in Figure \ref{data_synthesis}, where non-sensitive features are visualized using t-SNE \cite{tsne}. We further detail the data synthesis process and provide an overview of the synthetic data (divided by the minor sensitive attribute) in Appendix \ref{synthesis}. As shown in these figures, even after the removal of sensitive attributes, significant differences in feature and topology distributions persist between the two demographic subgroups, indicating the presence of both feature and topology biases. 


\subsection{Case Analysis}

\subsubsection{\textbf{Sensitive Correlation}} To unify the standard pipeline, we first measure $\mathcal{R}^2_{n}(\mathcal{X}, S)$, $\mathcal{R}^2_{n}(\mathcal{X}_{N}, S)$ and $h_{sens}$ for $S_n$ and $S_m$, which denote sensitive distance correlations of original features and non-sensitive features, and sensitive homophily of $S_n$ and $S_m$, respectively. We next obtain the prediction $\hat{Y}$, compute $\mathcal{R}^2_{n}(\hat{Y}, S)$, record $\Delta$SP and $\Delta$EO for $S_n$ and $S_m$, and compare them to evaluate biases. The split ratio is 1:1:8 for training, validation and test sets and we repeat the experiments 10 times to report the average results. The experimental setting is detailed in Appendix \ref{prelim}. 

\begin{table*}[h]
\caption{Sensitive Correlation in Before/After GNN Training. The results are shown in percentage($\%$). 1 represents the minor sensitive attribute while 2 denotes the major.}
\label{sens_cor_four_case}
\resizebox{\linewidth}{!}{%
\begin{tabular}{cccccccccc}
\hline

\multirow{2}{*}{\textbf{Cases}}     &  \multicolumn{4}{c}{\textbf{Before Training}}  & \multicolumn{5}{c}{\textbf{After Training}} \\ 
\cmidrule(lr){2-5} 
\cmidrule(lr){6-10} 
            &\textbf{$\boldsymbol{\mathcal{R}^2_{n}(\mathcal{X}, S)}$}  & \textbf{$\boldsymbol{\mathcal{R}^2_{n}(\mathcal{X}_{N}, S)}$}  &
            \textbf{$\boldsymbol{h_{sens1}}$} &
            \textbf{$\boldsymbol{h_{sens2}}$} &
            \textbf{\bm{$\Delta SP_1$}}& 
            \textbf{\bm{$\Delta EO_1$}} &
            \textbf{\bm{$\Delta SP_2$}} & 
            \textbf{\bm{$\Delta EO_2$}} &
            \textbf{$\boldsymbol{\mathcal{R}^2_{n}(\hat{Y}, S)}$}\\ \hline
            \textbf{BFDT}
                           & 72.76 
                           & 72.68
                           & 56.52
                           & 65.22
                           & 29.17$\pm 1.7$ 
                           & 36.20$\pm 1.8$ 
                           & 15.43$\pm 1.8$ 
                           & \textbf{8.87$\pm 2.0$} 
                           & 26.76$\pm 1.1$\\ \hline
            \textbf{DFBT}
                           & 29.50 
                           & 4.46 
                           & 66.89
                           & 79.70
                           & 11.91$\pm 0.6$ 
                           & 17.69$\pm 0.7$ 
                           & 35.94$\pm 0.9$ 
                           & 42.82$\pm 0.7$ 
                           & 16.98$\pm 0.1$\\ \hline
            \textbf{BFBT}
                           & 72.76
                           & 72.68
                           & 66.89 
                           & 79.70 
                           & 11.62$\pm 0.9$ 
                           & \textbf{0.60$\pm 0.3$} 
                           & 35.82$\pm 1.6$ 
                           & 18.73$\pm 1.3$ 
                           & 27.51$\pm 2.1$\\ \hline
            \textbf{DFDT}
                           & 29.50 
                           & 4.46 
                           & 56.53
                           & 65.22
                           & \textbf{6.02$\pm 2.3$} 
                           & 8.42$\pm 3.1$ 
                           & \textbf{10.46$\pm 4.2$} 
                           & 12.59$\pm 5.6$ 
                           & \textbf{12.57$\pm 0.3$}
                           \\ \hline
\end{tabular}
}
\end{table*}

\textbf{Case 1: Biased Features and Debiased Topology (BFDT)} In Case 1, we feed biased non-sensitive features and debiased topology into GNNs. As shown in Table \ref{sens_cor_four_case}, in terms of feature biases, the original sensitive $dcor$ is $72.76\%$. After removing two sensitive attributes, the sensitive $dcor$ decreases by only $0.08\%$, indicating small difference before and after masking sensitive attributes. As for the debiased graph topology, the sensitive homophily values of two sensitive attributes are relatively lower. After MP, biases from two sensitive attributes are projected into the predicted results, resulting in higher inequalities among different subgroups. The higher sensitive $dcor$ between the final prediction and two sensitive attributes also supports this finding. 




\textbf{Case 2: Debiased Features and Biased Topology (DFBT)} In Case 2, we focus on debiased features and biased topology. From Table \ref{sens_cor_four_case}, there is a large difference before and after masking the sensitive attributes, as the sensitive $dcor$ decreases by $25.04\%$. The sensitive homophily values of two sensitive attributes are relatively higher, suggesting the existence of biased topological structures. Interestingly, when utilizing fairness metrics ($\Delta SP$ and $\Delta EO$) separately for each sensitive attribute, the performance of the major sensitive attribute in Case 2 is not fairer than in Case 1, while the performance of the minor sensitive attribute largely improves. However, when using $dCor$ to handle the multiple sensitive attributes simultaneously, surpasses that of Case 1.

\textbf{Case 3: Biased Feature and Topology (BFBT)} In Case 3, we shift to biased non-sensitive features and topology. As shown in Table \ref{sens_cor_four_case}, similar to Case 1, higher sensitive $dcor$s are introduced. And similar to Case 2, the sensitive homophily values are relatively higher. We observe that the fairness performance of the minor sensitive attribute is much better than the major, although the major's performance excels the Case 2. However, when biases for both sensitive attributes are quantified simultaneously using sensitive $dcor$, the final performance is more biased than in Case 1, aligning with the intuition that biased features and topologies exacerbate group inequalities. 

\textbf{Case 4: Debiased Feature and Topology (DFDT)} In Case 4, we turn to debiased features and topology. In Table \ref{sens_cor_four_case}, when measured by $\Delta SP$, this case shows the best fairness performance compared to the other cases, and the performance in $\Delta EO$ is also relatively better, which indicates the fairness gaps between multiple sensitive attributes are further bridged. Despite using debiased features and topology, relatively higher group inequalities persist, which to some degree supports prior findings \cite{dai2021sayno_fair} that MP can exacerbate biases hidden behind node features. The fairness performance quantified by the sensitive $dcor$ also supports these findings. We notice that the sensitive $dcor$ in Case 1 is 14.19\% higher than in Case 4, attributed to biased node features, while in Case 2, the sensitive $dcor$ increases by 4.14\%, due to biased topology. This demonstrates that the majority of biases stem from biased node features, while graph topologies and MP mainly play complementary roles in amplifying biases.

\textbf{Discussion} The case analyses elucidate that node features, graph topologies and MP are all crucial bias sources, which motivates us to simultaneously debias features and topologies under MP at the pre-processing stage instead of debiasing them separately without taking MP into consideration. Moreover, using classical fairness metrics to treat multiple sensitive attributes separately and directly incorporate each one into objective functions as many fairness studies \cite{fair_reg, fair_reg2} did may not be a good choice. We argue that $dCor$ is more efficient to handle the complex relationships among multiple sensitive attributes. 
\subsubsection{\textbf{Sensitive Information Leakage}}
Our work considers a universal situation where due to fair awareness or legal compliance, the data owners (e.g., research institutions or companies) release masked non-sensitive features and incomplete graph topologies to the public for specific purposes. 
We argue that even though the usual procedures pre-handle features and topologies that are ready for release, the sensitive information leakage problem still exists and sensitive attributes can be inferred from the aforementioned resources, which further amplify existing inequalities and cause severe social consequences.
We assume that the adversaries can access pre-processed $\tilde{\mathcal{X}},\tilde{A}$ and label $\mathcal{Y}$, and then obtain partial sensitive attributes $S_p$ of specific individuals as prior knowledge, where $p\in\{0.3125\%,0.625\%,1.25\%,2.5\%,5\%,10\%,20\%\}$. Finally, they simply use a 1-layer GCN with a linear classifier as the attack model to identify different sensitive memberships, i.e., $S_m$ or $S_n$ of the rest nodes. We adopt the synthetic data again to explore sensitive information leakage with or without fairness intervention. 

\begin{figure}[htbp]
    \centering
    \subfloat[Attack Accuracy]{
    \includegraphics[width=0.45\textwidth]{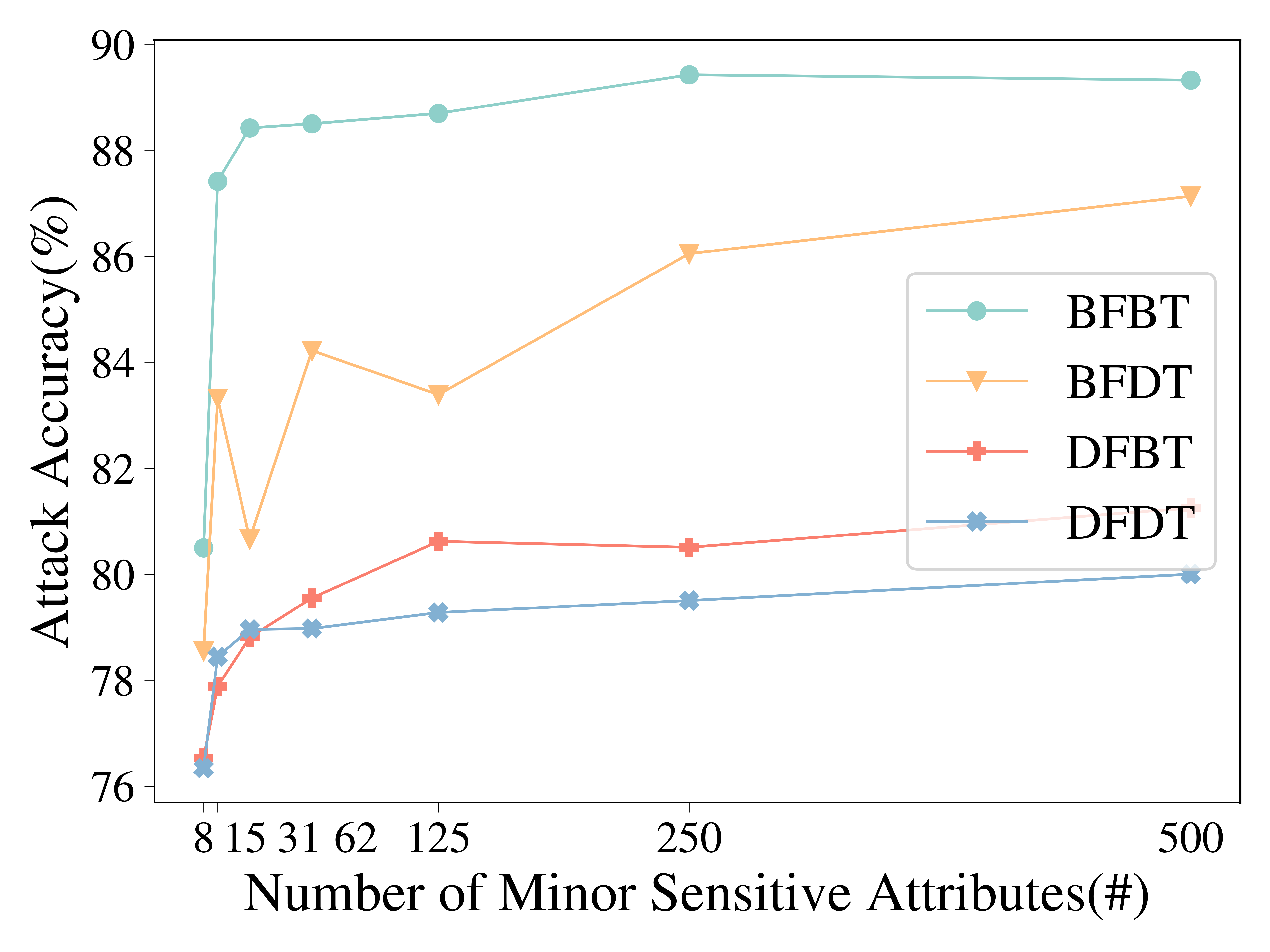}
    \includegraphics[width=0.45\textwidth]{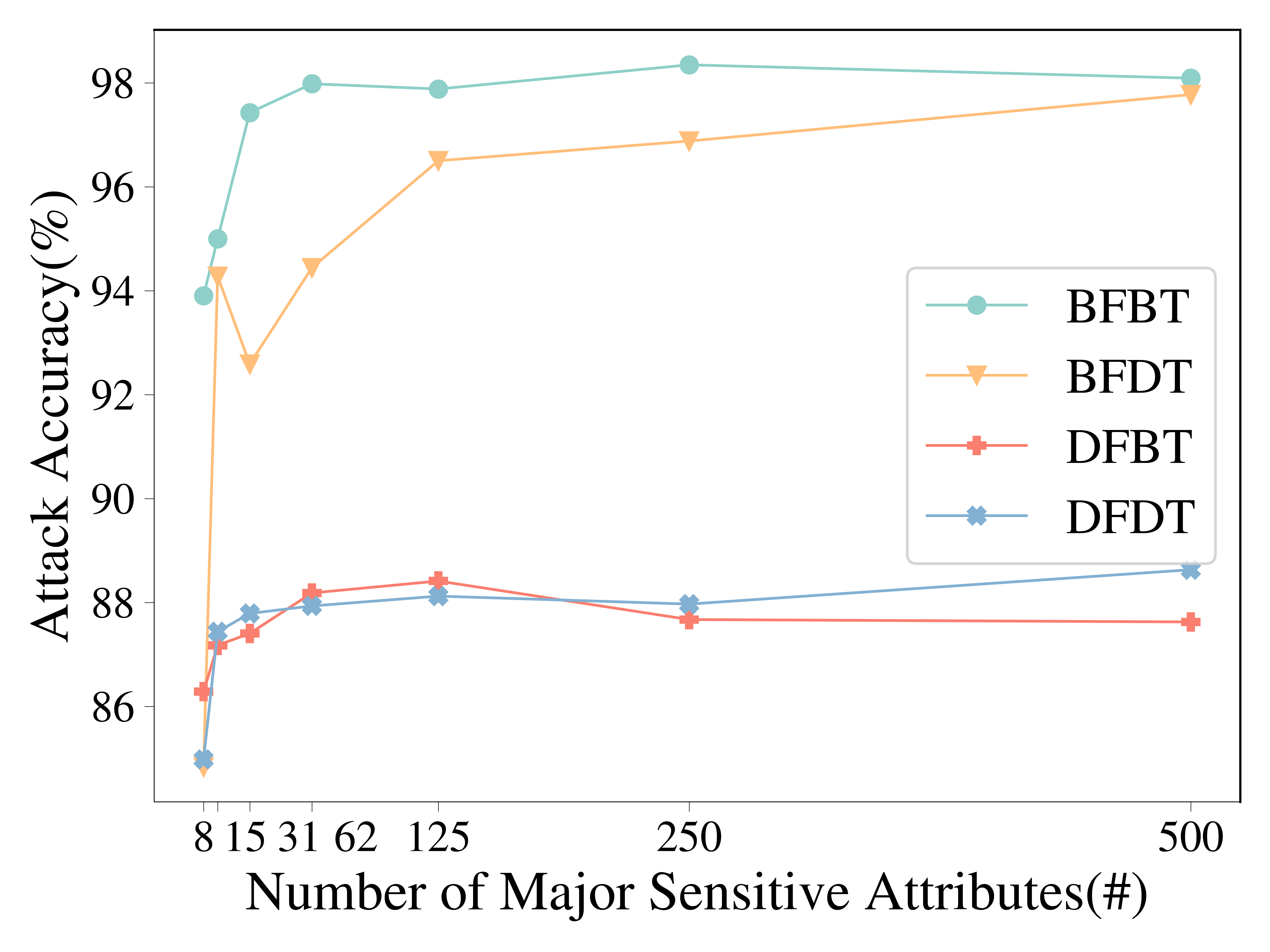}
    \label{acc_major}}
    \hfill
    \subfloat[Sensitive Correlation]{
    \includegraphics[width=0.45\textwidth]{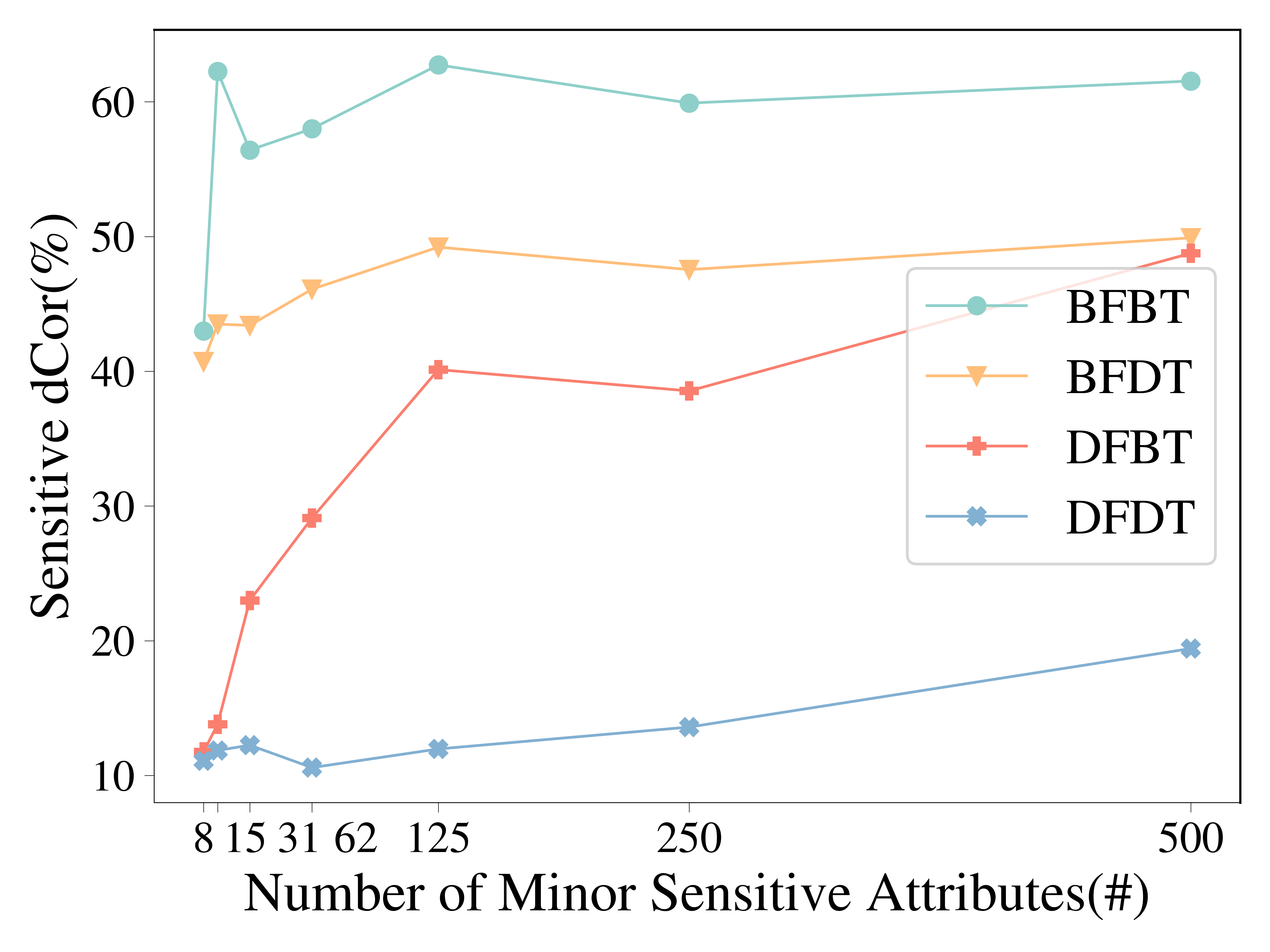}
    \includegraphics[width=0.45\textwidth]{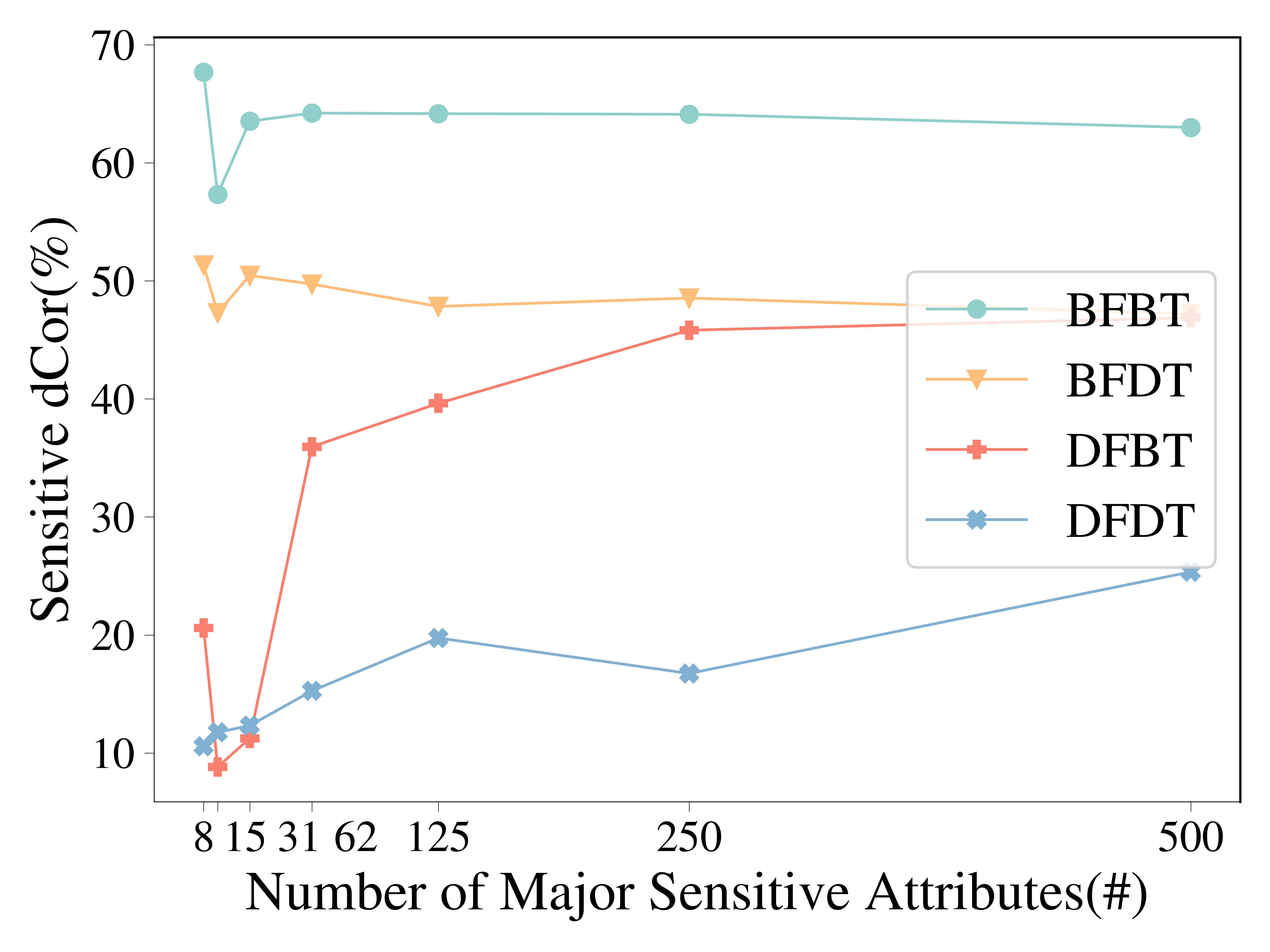}
    \label{dcor_major}}
    \caption{Attribute Inference Attack Under Different Cases}
    \label{fig:att_syn}
\end{figure}

As Figure \ref{fig:att_syn} shows, even though the adversaries only acquire very few $S_m$ or $S_n$, they can successfully infer the rest from all pairs since highly associated features are retained and become more sensitively correlated after MP. We note that $S_m$ are more biased than $S_n$. While there are turning points in the fewer label cases due to performance instability, as more sensitive labels are collected, both attack accuracy and sensitive correlations increase. Overall, the BFBT pair consistently introduces the most biases and sensitive information leakage, 
the BFDT pair leads to lower sensitive correlation and attack accuracy, and the performances of DFBT and DFDT pairs are close, which indicates that compared to biased features, biased topology contributes less to inference attacks of $S_m$ and $S_n$. Once fairness interventions for features and topology are introduced, the overall attack accuracy stabilizes to decrease by 10$\%$ and the sensitive correlation decreases by almost 50$\%$/40$\%$ for $S_m$ and $S_n$. 

\textbf{Discussion} The above analysis illustrates that even simple fairness interventions can alleviate attribute inference attacks under the black-box settings. Generally, more advanced debiasing methods will result in less sensitive information leakage, which fits our intuition and motivates us to design more effective debiasing techniques with limited sensitive information leakage.

\subsection{Problem Statement}
Based on the two empirical studies, we define the formal problem as: 
\textit{{Given an undirected attributed graph $\mathcal{G}=\left(\mathcal{V}, \mathcal{E}, \mathcal{X}\right)$ with the sensitive attributes $S$, non-sensitive features $\mathcal{X}_N$, graph topology $A$ and node labels $\mathcal{Y}$, we aim to learn pre-debiasing functions $\Phi_f(X)=\hat{X}$ and $\Phi_t(A)=\hat{A}$ and thereby construct a fair and model-agnostic classifier $f_{\theta}(\hat{X}, \hat{A})=\hat{Y}$ with limited sensitive information leakage.}}

%% file: Framework.tex
\section{Framework Design}
In this section, we first provide an overview of MAPPING, which sequentially debias node features and graph topologies, and then we detail each debiasing module to tackle the formulated problem.

\subsection{Framework Overview}

\begin{figure*}[!htbp]
    \centering
    \includegraphics[width=13cm]{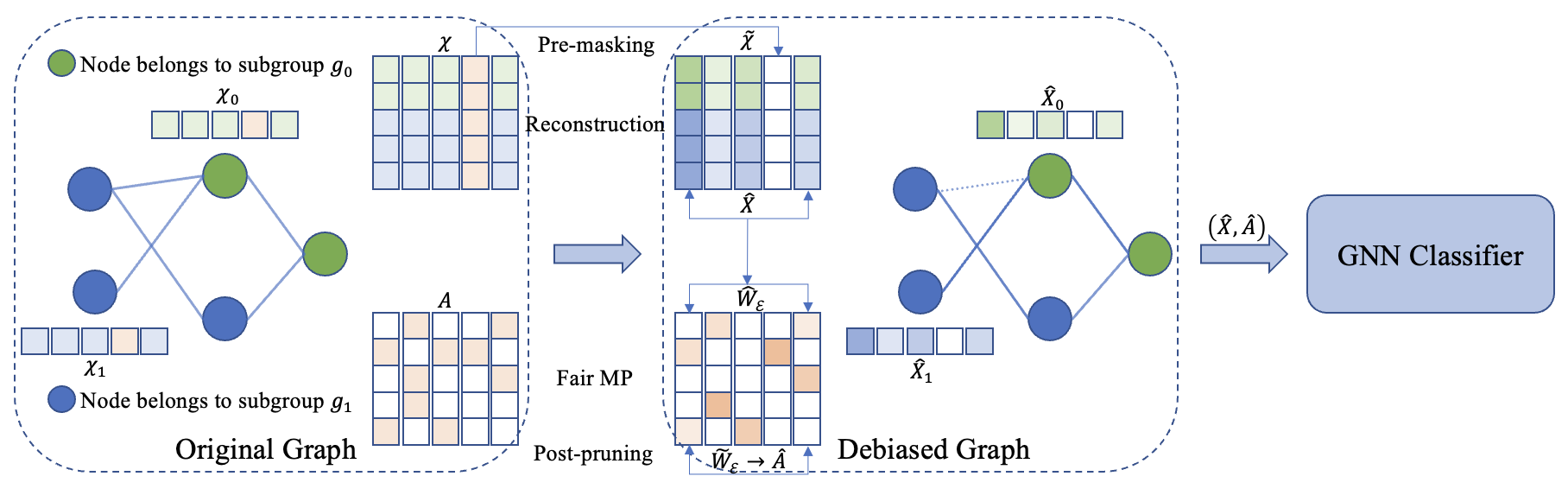}
    \caption{\textbf{The Framework Overview of MAPPING with Feature and Topology Debiasing Modules. The feature debiasing module contains Pre-masking and Reconstruction submodules and the Topology debiasing module includes Fair MP and Post-pruning submodules. We implement these two modules sequentially.}}
    \label{MAPPING_overview}
 \end{figure*}
\vspace{-1mm}

MAPPING consists of two modules to debias node features and graph topologies. The feature debiasing module contains $1)$ pre-masking: masking sensitive attributes and their highly associated features based on hard rules to trade off utility and fairness and $2)$ reconstruction: reconstructing the pre-masked features $\Tilde{\mathcal{X}}$ to restrict attribute inference attacks by adversarial training and $dCov$-based fairness constraints. With debiased features $\hat{X}$ on hand, the topology debiasing module includes $1)$ Fair MP: initializing equalized weights $W_0$ for existing edges $\mathcal{E}$ and then employing the $dCov$-based fairness constraint to mitigate sensitive privacy leakage and
$2)$ Post-pruning: pruning edges with weights $\hat{W}_{\mathcal{E}}$ beyond the pruning threshold $r_p$ to obtain $\Tilde{W}_{\mathcal{E}}$ and then returning the debiased adjacency matrix $\hat{A}$. The overview of MAPPING is shown in Figure \ref{MAPPING_overview}. And the algorithm overview is detailed in Algorithm \ref{alg2} in Appendix \ref{code}.

\subsection{Feature Debiasing Module}

\subsubsection{\textbf{Pre-masking}}

Prior fairness studies commonly remove all sensitive attributes before GNN training. However, simple removal, i.e., fairness through blindness \cite{fairthroughaware}, cannot sufficiently protect specific demographic groups from discrimination and attacks. $\mathcal{X}_{N}$ that are highly associated with $S$ can reveal the sensitive membership or be used to infer $S$ even without access to it \cite{fairthroughaware, main_attr_infer_attack}. Furthermore, 
without fairness intervention, $\mathcal{Y}$ are always reflective of societal inequalities and stereotypes \cite{societal_bias_label} and $\hat{Y}$ ineluctably inherit such biases since they are predicted by minimizing the difference between $\mathcal{Y}$ and $\hat{Y}$. Inspired by these points, we first design a pre-masking method to mask $S$ and highly associated $\mathcal{X}_N$ with the power of $dCor$. We cannot simply discard all highly associated $\mathcal{X}_N$ since part of them may carry useful information and finally contribute to node classification. Hence, we must carefully make the trade-off between fairness and utility. Please note that Pre-masking only provides a coarse screening, more feature biases will be mitigated in the Reconstruction submodule.

Considering above factors, we first compute $\mathcal{R}_{n}^{2}(\mathcal{X}_i, S)$ and $\mathcal{R}_{n}^{2}( \mathcal{X}_i, \mathcal{Y})$, $i\in[1,\dots,d]$. We set a distributed ratio $r$ (e.g., $20\%$) to pick up $x$ top related features based on $\mathcal{R}_{n}^{2}(\mathcal{X}, S)$ and $x$ less related features based on $\mathcal{R}_{n}^{2}(\mathcal{X}, \mathcal{Y})$. We then take an intersection of these two sets to acquire features that are highly associated with $S$ and simultaneously contribute less to accurate node classification. Next, we use a sensitive threshold $r_s$ (e.g., $70\%$) to filter features whose $\mathcal{R}_{n}^{2}(\mathcal{X}_i, S)<r_s$. Finally, we take the union of these two sets of features to guarantee: 
\begin{itemize}
    \item \textit{We cut off very highly associated $\mathcal{X}_{N}$ to pursue fairness;}
    \item \textit{Besides the hard rule, to promise accuracy, we scrutinize that only $\mathcal{X}$ are highly associated with $S$ and make fewer contributions to the final prediction are masked;}
    \item \textit{The rest of the features are pre-masked features $\Tilde{\mathcal{X}}$.}
\end{itemize}
More details are seen in Algorithm \ref{alg1} in Appendix \ref{code}.

Pre-masking benefits bias mitigation and privacy preservation. Besides, it reduces the dimension of node features, thereby saving training costs, which is particularly effective on large-scale datasets with higher dimensions. However, partially masking $S$ and its highly associated $\mathcal{X}_{N}$ may not be adequate. Prior studies \cite{fairclasswithoutsens,without_sens_learning} have demonstrated the feasibility of $S$ estimation without accessing $S$. To tackle this issue, we further debias node features after pre-masking. Please note that $r$ and $r_s$ are all experienced values, if $r$ is too large and $r_s$ is too small, more related features will be directly removed, which may hurt accuracy. We recommend conservative choices for these two values since more biases can be mitigated in the next submodule. 

\subsubsection{\textbf{Reconstruction}} 
 We first assign initially equal weights $W_{f_{0}}$ for $\Tilde{\mathcal{X}}$. For feature $\Tilde{\mathcal{X}_i},i\in[1,\dots, d_m]$, where $d_m$ is the dimension of masked features, if corresponding $\hat{W}_{fi}$ decreases, $\Tilde{\mathcal{X}_i}$ plays a less important role to debias features and vice versa. Therefore, the first objective is to minimize: 
 \begin{equation}
    \setlength{\abovedisplayskip}{3pt}
    \setlength{\belowdisplayskip}{3pt}
     \min_{\theta_{r}}\mathcal{L}_r=\|\Tilde{\mathcal{X}}-\hat{X}\|_{2}
 \end{equation}
where $\hat{X}=f_{\theta_{\hat{w}}}(\Tilde{\mathcal{X}})=\Tilde{\mathcal{X}}\hat{W}_f$.

In addition, we restrict the weight changes and control the sparsity of weights. Here we introduce $L$1 regularization as: 
\begin{equation}
    \setlength{\abovedisplayskip}{3pt}
    \setlength{\belowdisplayskip}{3pt}
    \min_{\theta_{\hat{w}}}\mathcal{L}_{\hat{w}}=\|\hat{W}_f\|_{1}
\end{equation}

Next, we introduce $dCov$-based fairness constraints. Since the same technique is utilized in fair MP, we unify them to avoid repeated discussions. The ideal cases are $\hat{X} \perp S$ and $\hat{Y} \perp S$, which indicates that the sensitive attribute inference derived from $\hat{X}$ and $\hat{Y}$ are close to random guessing.  Zafar et al. \cite{fairtreat17} and Dai et al. \cite{dai2021sayno_fair} employ $Cov$-based constraints to learn fair classifiers. However, $Cov$ needs pairwise computation, and it ranges from $-\infty$ to $\infty$, which requires adding the extra absolute form. Moreover, $Cov=0$ cannot ensure independence but only reflect irrelevance. Cho et al. \cite{fairmutual} and Roh et al. \cite{fairmutualtrain} use $MI$-based constraints for fair classification. Though $MI$ uncovers mutual independence between random variables, it cannot get rid of dimensional restrictions. Whereas $dCov$ can overcome these deficiencies. $dCov$ reveals independence and it is larger than $0$ for dependent cases. Above all, it breaks dimensional limits and thereby saves computation costs. Hence, we leverage a $dCov$-based fairness constraint in our optimization, marked as: 
\begin{equation}
    \setlength{\abovedisplayskip}{3pt}
    \setlength{\belowdisplayskip}{3pt}
    \mathcal{L}_{s}=\mathcal{V}_{n}^2(\hat{X}, S)
\end{equation}

Finally, we use adversarial training \cite{fair_ad_learn} to mitigate sensitive privacy leakage by maximizing the classification loss as: 
\begin{equation}
    \label{sens}
    \setlength{\abovedisplayskip}{3pt}
    \setlength{\belowdisplayskip}{3pt}
    \max_{\theta_{a}}\mathcal{L}_{a}=-\frac{1}{n}\sum_{i=1}^{n}S_i\log(\hat{S}_i)+(1-S_i)\log(1-\hat{S}_i)
\end{equation}
where $\hat{S} = f_{\theta_{{s}}}(\hat{X})$.

Cho et al. \cite{fairmutual} empirically shows that adversarial training suffers from significant stability issues and $Cov$-based constraints are commonly adopted to alleviate such instability. In this paper, we use $dCov$ to achieve the same goal. 

\subsubsection{\textbf{Final Objective Function of Feature Debiasing}}
Now we have $f_{\theta_{r}}$ to minimize feature reconstruction loss, $f_{\theta_{\hat{w}}}$ to restrict weights, $f_{\theta_{a}}$ to debias adversarially, and $\mathcal{L}_{s}$ to diminish sensitive information disclosure. In summary, the final objective function of the feature debiasing module is: 
\begin{equation}
    \setlength{\abovedisplayskip}{3pt}
    \setlength{\belowdisplayskip}{3pt}\min_{\theta_{r},\theta_{\hat{w}},\theta_{a}}\mathcal{L}_r+\lambda_1 \mathcal{L}_{\hat{w}} + \lambda_2 \mathcal{L}_s - \lambda_3 \mathcal{L}_a
\end{equation}
where $\theta_r$, $\theta_{\hat{w}}$ and $\theta_a$ denote corresponding objective functions' parameters. Coefficients $\lambda_1$, $\lambda_2$ and $\lambda_3$ control weight sparsity, sensitive correlation and adversarial debiasing, respectively.

\subsection{Topology Debiasing Module}

\subsubsection{\textbf{Fair MP}}

Solely debiasing node features is not
sufficient, prior work \cite{dai2021sayno_fair} empirically demonstrates biases can be magnified by graph topologies and MP. 
FMP \cite{fmp_topology_bias} investigates how topologies can enhance biases during MP. Our empirical analysis likewise reveals that sensitive correlations increase after MP. 
Here we propose a novel debiasing method to jointly debias topologies under MP, which is conducive to providing post-pruning with explanations of edge importance.

First, we initialize equal weights $W_{{\mathcal{E}_0}}$ for $\mathcal{E}$, then feed $W_{{\mathcal{E}}_0}$, $A$ and $\hat{X}$ into GNN training. The main goal of node classification is to pursue accuracy, which equalizes to minimize:
\begin{equation}
    \setlength{\abovedisplayskip}{3pt}
    \setlength{\belowdisplayskip}{3pt}
    \min_{\theta_{\mathcal{C}}}\mathcal{L}_{\mathcal{C}}=-\frac{1}{n}\sum_{i=1}^{n}\mathcal{Y}_i\log(\hat{Y}_i)+(1-\mathcal{Y}_i)\log(1-\hat{Y}_i)
\end{equation}

As mentioned before, we add a $dCov$-based fairness constraint into the objective function to ameliorate sensitive attribute inference attacks derived from $\hat{Y}$. Defined as:
\begin{equation}
    \setlength{\abovedisplayskip}{3pt}
    \setlength{\belowdisplayskip}{3pt}
    \mathcal{L}_{\mathcal{F}}=\mathcal{V}^2_{n}(\hat{Y},S)
\end{equation}

\vspace{-1mm}
\subsubsection{\textbf{Final Objective Function of Topology Debiasing}}
Now we have $f_{\theta_{\mathcal{C}}}$ to minimize node classification loss and $\mathcal{L}_{\mathcal{F}}$ to restrain sensitive information leakage. The final objective function of the topology debiasing module can be written as:
\begin{equation}
    \setlength{\abovedisplayskip}{3pt}
    \setlength{\belowdisplayskip}{3pt}
    \min_{f_{\theta_{\mathcal{C}}}}\mathcal{L}_e = \mathcal{L}_{\mathcal{C}} + \lambda_4\mathcal{L}_{\mathcal{F}}
\end{equation}
Where $\theta_{\mathcal{C}}$ denotes the parameter of the node classifier $\mathcal{C}$, the coefficient $\lambda_4$ controls the balance between utility and fairness.

\subsubsection{\textbf{Post-pruning}}
After learnable weights $\hat{W}_{\mathcal{E}}$ have been updated to construct a fair node classifier with limited sensitive information leakage, we apply a hard rule to prune edges with edge weights $\hat{W}_{e_i\cdot}$ for $e_{i\cdot}\in\mathcal{E}$ that are beyond the pruning threshold $r_p$. Please note, since we target undirected graphs, if any two nodes $i$ and $j$ are connected, $\hat{W}_{e_{ij}}$ should be equal to $\hat{W}_{e_{ji}}$, which assures $A_{ij}=A_{ji}$ after pruning. Mathematically,
\begin{equation}
    \setlength{\abovedisplayskip}{3pt}
    \setlength{\belowdisplayskip}{3pt}
    \begin{aligned}
        & \Tilde{W}_{ei}=
\begin{cases}
    1, & \mbox{if }\hat{W}_{e_{i\cdot}} \ge r_p\\
0, & \mbox{otherwise} 
\end{cases} \\
    &  s.t.,\hat{W}_{e_{ij}}=\hat{W}_{e_{ji}}, \mbox{ if } e_{ij}\in\mathcal{E}
    \end{aligned}
\end{equation}

Removing edges is practical since edges can be noisy in reality, and meanwhile, they can exacerbate biases and leak privacy under MP. $\hat{W}_{e_i}$ explains which edges contribute less/more to fair node classification. We simply discard identified uninformative and biased edges. The rest forms the new $\hat{A}$.

\vspace{-2mm}
\subsection{Extension to Multiple Sensitive Attributes} 
Since $dCor$ and $dCov$ are not limited by dimensions, our work can be easily extended into the pre-masking submodule and adding $dCov$-based fairness constraints $\mathcal{L}_s$ and $\mathcal{L}_{\mathcal{F}}$ with multiple sensitive attributes as we directly calculate in previous sections. As for adversarial training, the binary classification loss cannot be trivially extended to multiple sensitive labels. Previously in the Reconstruction submodule, we simply adopted $f_{\theta_s}$ to map estimated masked features with dimension $d_s$ into a single predicted sensitive attribute. Instead, in the multiple case, we map the features into predicted multiple sensitive attributes with the desired dimension $d_m$. Next, we leverage the function (\ref{sens}) to compute the classification loss for each sensitive attribute and then take the average. The rest steps are exactly the same as described before.



%% file: Experiments.tex
\section{Experiments}
In this section, we implement a series of experiments to demonstrate the effectiveness and flexibility of MAPPING with different GNN variants. Particularly, we address the following questions: 
\begin{itemize}
    \item \textbf{Q1:} Does MAPPING effectively and efficiently debias feature and topology biases hidden behind graphs? 
    \item \textbf{Q2:} Does MAPPING flexibly adapt to different GNNs?
    \item \textbf{Q3:} Does MAPPING outperform existing pre-processing and in-processing algorithms for fair node classification?
    \item \textbf{Q4:} Whether MAPPING can achieve better trade-offs between utility and fairness and meanwhile mitigate sensitive information leakage?
    \item \textbf{Q5:} How debiasing contribute to fair node classification?
\end{itemize}

\subsection{Experimental Setup}
In this subsection, we first describe the datasets, metrics and baselines, and then summarize the implementation details.

\subsubsection{\textbf{Datasets}}
We validate MAPPING on three real-world datasets, namely, German, Recidivism and Credit \cite{nifty}\footnote{https://github.com/chirag126/nifty}. 
The detailed statistics are summarized in Table \ref{data_summary} and more contents are in Appendix \ref{data}. 
\begin{table*}[htbp]
 \caption{\textbf{Statistics Summary of Datasets}}
 \resizebox{\linewidth}{!}{
  \centering
  \begin{tabular}{llll}
    \toprule
    \textbf{Dataset}  & \textbf{German} & \textbf{Recidivism} & \textbf{Credit} \\
    \midrule
    \textbf{\# Nodes}  & 1000 & 18,876 & 30,000    \\
    \textbf{\# Edges}  & 22,242 & 321,308 & 1,436,858     \\
    \textbf{\# Features}  & 27 & 18 & 13  \\
    \textbf{Sensitive Attr.}  & Gender(Male/female) & Race(Black/white) & Age($\le 25$/$>25$)  \\
    \textbf{Label} & Good/bad credit & Bail/no bail  & Default/no default  \\
    \bottomrule
  \end{tabular}}
  \label{data_summary}
\end{table*}

\subsubsection{\textbf{Evaluation Metrics}} 
We adopt accuracy (ACC), F1 and AUROC to evaluate utility and $\Delta_{SP}$ and $\Delta_{EO}$ to measure fairness.  

\subsubsection{\textbf{Baselines}} 
We investigate the effectiveness and flexibility of MAPPING on three representative GNNs, namely, GCN, GraphSAGE \cite{GraphSAGE} and GIN \cite{GIN}, and compare MAPPING with three state-of-the-art debiasing models.

\textbf{Vanilla} 
GCN leverages a convolutional aggregator to sum propagated features from local neighborhoods. GraphSAGE aggregates node features from local sampled neighbors, which is more scalable to handle unseen nodes. GIN emphasizes the expressive power of graph-level representation to satisfy the Weisfeiler-Lehman graph isomorphism test \cite{WL_test}. Moreover, these three GNNs adopt diverse MP, which is conducive to investigating debiasing effects of the Fair MP submodule of MAPPING under different MP.  

\textbf{State-of-the-art (SOTA) Debiasing Models} 
We choose one pre-processing model, namely EDITS \cite{dong2022edits} and two in-processing models, namely, FairGNN \cite{dai2021sayno_fair} and NIFTY \cite{nifty}. 
EDITS proposes a model-agnostic debiasing framework based on Wasserstein distance, which reduces feature and structural biases by feature retuning and edge clipping. FairGNN predicts missing sensitive attributes via a sensitive attribute estimator and sequentially combines adversarial debiasing and covariance constraints to learn a fair GNN classifier. NIFTY endeavors to learn a fair and stable node representation under counterfactual perturbation. Since NIFTY \cite{nifty} targets fair node representation, we evaluate the quality of node representation on the downstream node classification task. 

\subsubsection{\textbf{Implementation Details}}
We keep the same experiment setting as before. The GNNs follow the same architectures in NIFTY \cite{nifty}. The fine-tuning processes are handled with Optuna \cite{optuna} via the grid search. As for feature debiasing, we deploy a $1$-layer multilayer perceptron (MLP) for adversarial debiasing and leverage the proximal gradient descent method to optimize $W_f$. Since PyTorch Geometric only allows positive weights, we use a simple sigmoid function to transfer weights into $[0,1]$. We set the learning rate as $0.001$ and the weight decay as 1e-5 for all three datasets, and set training epochs as $500$. For topology debiasing, we adopt a $1$-layer GCN to mitigate biases under MP. We set training epochs as $1000$ for all datasets, as for GIN in Credit, since small epochs can achieve comparable performance, we set early stopping to avoid overfitting. Others are the same as feature debiasing. As for GNN training, we utilize the split setting in NIFTY \cite{nifty}, we fix the hidden layer as $16$, the dropout as $0.2$, training epochs as $1000$, weight decay as 1e-5 and learning rate from $\{0.01,0.03\}$ for all GNNs. For fair comparison, we rigorously follow the settings in SOTA \cite{dai2021sayno_fair,nifty,dong2022edits}. The other detailed hyperparameter settings are in Appendix \ref{setting}.

The attack setting is the same as before to evaluate sensitive information leakage. Still, we repeat experiments $10$ times with $10$ different seeds and finally report the average results. All experiments are conducted on a 64-bit machine with 4 Nvidia A100 GPUs.


\subsection{Performance Evaluation}
\label{perf}
In this subsection, we evaluate the performance by addressing the top $4$ questions raised at the beginning of this section.

\subsubsection{\textbf{Debiasing Effectiveness and Efficiency}}
To answer \textbf{Q1}, we first compute $\Delta$SP and $\Delta$EO before and after debiasing and then evaluate the debiasing effectiveness. Second, we provide the time complexity analysis to illustrate the efficiency of MAPPING.
\begin{table*}[h]
\caption{\textbf{Node classification performance comparison on German, Recidivism and Credit.}}
\resizebox{\linewidth}{!}{
\begin{tabular}{ccccccccccccccccc}
\hline

\multirow{2}{*}{\textbf{GNN}}       & \multirow{2}{*}{\textbf{Framework}} & \multicolumn{5}{c}{\textbf{German}}  & \multicolumn{5}{c}{\textbf{Recidivism}} & \multicolumn{5}{c}{\textbf{Credit}} \\ 
\cmidrule(lr){3-7} 
\cmidrule(lr){8-12} 
\cmidrule(lr){13-17} 
                           &                            & \textbf{ACC} & \textbf{F1}  & \textbf{AUC} & \textbf{\bm{$\Delta$}SP} & \textbf{\bm{$\Delta$}EO} 
                           %
                           &           \textbf{ACC} & \textbf{F1}  & \textbf{AUC} & \textbf{\bm{$\Delta$}SP} & \textbf{\bm{$\Delta$}EO} 
                           %
                           & \textbf{ACC} & \textbf{F1}  & \textbf{AUC} & \textbf{\bm{$\Delta$}SP} & \textbf{\bm{$\Delta$}EO} 
                           \\ \hline
\multirow{5}{*}{\textbf{GCN}} 
            & Vanilla                    
            & \textbf{72.00}$\pm 2.8$     & 80.27$\pm 2.5$    & \textbf{74.34}$\pm 2.4$    &  31.42$\pm 9.5$  & 22.56$\pm 6.2$    
            & 87.54$\pm 0.1$    &  82.51$\pm 0.1$  &  91.11$\pm 0.1$   & 9.28$\pm 0.1$   & 8.19$\pm 0.3$   
            & 76.19$\pm 0.4$     & 84.22$\pm 0.4$    & \textbf{73.34}$\pm 0.0$    & 9.00$\pm 1.2$   & 6.12$\pm 0.9$  
            \\
            
            & FairGNN                      
            &  67.80$\pm 11.0$    &  74.10$\pm 17.6$   & 73.08$\pm 2.0$    &  24.37$\pm 8.7$  &  16.99$\pm 6.8$   
            & 87.50$\pm 0.2$    & 83.40$\pm 0.2$   &  91.53$\pm 0.1$   & 9.17$\pm 0.2$   &  7.93$\pm 0.4$  
            &  73.78$\pm 0.1$    &  82.01$\pm 0.0$   & 73.28$\pm 0.0$    & 12.29$\pm 0.6$   & 10.04$\pm 0.7$   \\
                           
            & NIFTY                      
            & 66.68$\pm 8.6$     & 73.59$\pm 13.4$    & 70.59$\pm 4.8$    &  15.65$\pm 9.2$  & 10.58$\pm 7.3$    
            & 76.67$\pm 1.8$    & 69.09$\pm 0.9$   & 81.27$\pm 0.4$    & 3.11$\pm 0.4$   &  2.78$\pm 0.5$  
            &  73.33$\pm 0.1$    &  81.62$\pm 0.1$   &  72.08$\pm 0.1$   &  11.63$\pm 0.2$  & 9.32$\pm 0.2$   \\
            
            & EDITS                    
            &  69.80$\pm 3.2$     &  80.18$\pm 2.2$   & 67.57$\pm 6.0$    & 4.85$\pm 2.8$   &  4.93$\pm 2.2$  
            & 84.82$\pm 0.8$    & 78.56$\pm 1.1$   &  87.42$\pm 0.7$   &  7.23$\pm 0.3$  & 4.43$\pm 0.7$  
            & 75.20$\pm 1.7$     & 84.11$\pm 2.2$    & 68.63$\pm 5.8$    &  5.33$\pm 3.8$  & 3.64$\pm 2.7$   
            \\
                           
            & MAPPING                
            & 70.84$\pm 1.8$     & \textbf{81.33}$\pm 1.3$    & 70.86$\pm 1.9$    &  \textbf{4.54}$\pm 2.2$  & \textbf{4.00}$\pm 1.7$   
            &  \textbf{88.91}$\pm 0.2$   &   \textbf{84.17}$\pm 0.4$ &  \textbf{93.31}$\pm 0.1$   &  \textbf{2.81}$\pm 0.2$  & \textbf{0.73}$\pm 0.3$   
            & \textbf{76.73}$\pm 0.2$     & \textbf{84.81}$\pm 0.2$    &  73.26$\pm 0.0$   & \textbf{1.39}$\pm 0.4$   & \textbf{0.21}$\pm 0.2$   
            \\ \hline
            
\multirow{5}{*}{\textbf{GraphSAGE}} 
            & Vanilla                    
            & 71.76$\pm 1.4$     & \textbf{81.86}$\pm 0.8$    & 71.10$\pm 3.1$    & 14.00$\pm 8.4$   & 7.10$\pm 4.9$   
            & 85.91$\pm 3.2$     & 81.20$\pm 2.9$   & 90.42$\pm 1.3$    & 4.42$\pm 3.1$   & 3.34$\pm 2.2$  
            & 78.68$\pm 0.8$     & 86.57$\pm 0.7$    &  74.22$\pm 0.4$   & 19.49$\pm 5.8$   &  15.92$\pm 5.5$  
            \\
            
            & FairGNN                      
            &  \textbf{73.80}$\pm 1.4$    & 81.13$\pm 1.1$    & \textbf{74.37}$\pm 1.0$    & 20.94$\pm 4.0$   &  12.05$\pm 3.8$   
            & \textbf{87.83}$\pm 1.0$    & \textbf{83.06}$\pm 1.1$   & 91.72$\pm 0.5$    &  3.73$\pm 1.9$  &  4.97$\pm 2.7$  
            &  72.99$\pm 1.9$    & 81.28$\pm 1.7$    & \textbf{75.60}$\pm 0.2$    &  11.63$\pm 4.9$  &  9.592$\pm 5.0$   
            \\
                           
            & NIFTY                      
            &  70.04$\pm 2.2$    & 78.77$\pm 2.5$    &  73.02$\pm 2.2$   & 16.93$\pm 8.0$   & 11.08$\pm 6.5$     
            & 84.53$\pm 6.5$    & 80.30$\pm 4.7$   & \textbf{91.99}$\pm 0.8$    & 5.92$\pm 1.2$   & 4.54$\pm 1.5$   
            & 73.64$\pm 1.5$     & 81.89$\pm 1.3$    & 73.33$\pm 0.2$   &  11.51$\pm 1.0$  & 9.05$\pm 0.9$  
            \\
            
            & EDITS                   
            & 69.76$\pm 1.5$     & 80.23$\pm 1.8$    &  69.35$\pm 1.5$   & 4.52$\pm 3.3$   & 6.03$\pm 5.5$ 
            & 85.13$\pm 1.1$    & 78.64$\pm 1.3$   &  89.36$\pm 0.9$   &  6.75$\pm 1.0$  &  5.14$\pm 1.2$  
            &  74.06$\pm 2.0$    & 82.34$\pm 1.8$   &  74.12$\pm 1.3$   & 13.00$\pm 8.6$   & 11.42$\pm 9.1$ 
            \\
            
            & MAPPING                
            & 70.76$\pm 1.2$     & 81.51$\pm 0.7$    & 69.89$\pm 1.9$    & \textbf{3.73}$\pm 3.1$   & \textbf{2.39}$\pm 1.5$   
            &  87.30$\pm 0.8$   & 82.07$\pm 1.2$   & 91.41$\pm 0.9$    & \textbf{3.54}$\pm 1.9$   &  \textbf{3.27}$\pm 1.8$  
            &  \textbf{80.19}$\pm 0.3$     &\textbf{88.24}$\pm 0.2$    & 74.07$\pm 0.6$    & \textbf{4.93}$\pm 0.8$   & \textbf{2.57}$\pm 0.6$   
            \\ \hline
            
\multirow{5}{*}{\textbf{GIN}}       
            & Vanilla                    
            & 71.88$\pm 1.5$     & 81.93$\pm 0.7$    & 67.21$\pm 10.3$    &  14.07$\pm 10.6$  & 9.78$\pm 8.2$    
            & \textbf{87.62}$\pm 3.7$     & \textbf{83.44}$\pm 4.5$  & \textbf{91.05}$\pm 3.1$    & 9.92$\pm 2.6$   & 7.75$\pm 2.1$   
            & 74.82$\pm 1.9$     & 83.31$\pm 2.1$   & 73.84$\pm 1.2$    & 9.40$\pm 4.5$   & 7.37$\pm 3.8$  
            \\
            
            & FairGNN                      
            &  65.32$\pm 10.4$    &  72.31$\pm 17.8$   &  66.07$\pm 8.7$   & 13.67$\pm 11.8$   &  10.91$\pm 11.2$ 
            & 84.32$\pm 1.8$    & 80.30$\pm 2.1$   &  90.19$\pm 1.3$   &  8.15$\pm 2.4$  & 6.28$\pm 1.4$ 
            &  72.22$\pm 0.5$    & 80.67$\pm 0.4$    & \textbf{74.87}$\pm 0.2$    &  12.52$\pm 3.2$  & 10.56$\pm 3.6$   
            \\
            
            & NIFTY                      
            & 64.96$\pm 5.9$     & 72.62$\pm 7.8$   &  67.70$\pm 4.1$   & 11.36$\pm 6.3$   &  10.07$\pm 6.9$ 
            &  83.52$\pm 1.6$   & 77.18$\pm 3.1$   & 87.56$\pm 0.9$    & 6.09$\pm 0.9$   & 5.65$\pm 1.1$  
            & 75.88$\pm 0.7$     & 83.96$\pm 0.8$    &  72.01$\pm 0.5$   & 11.36$\pm 1.8$   & 8.95$\pm 1.5$  
            \\
            
            & EDITS                    
            &  71.12$\pm 1.5$    &  81.63$\pm 1.3$   & 69.91$\pm 1.8$    &  3.04$\pm 2.6$  &  3.47$\pm 3.3$  
            & 75.73$\pm 7.8$    & 65.56$\pm 10.3$   &  77.57$\pm 9.1$   &  4.22$\pm 1.4$  & 3.35$\pm 1.2$  
            & 76.68$\pm 0.8$     & 85.15$\pm 0.7$    &  70.91$\pm 2.0$   &  5.52$\pm 3.9$  &  4.76$\pm 2.8$ 
            \\
            
            & MAPPING                
            & \textbf{73.40}$\pm 1.2$     & \textbf{83.18}$\pm 0.5$    & \textbf{71.48}$\pm 0.6$    & \textbf{2.20}$\pm 1.4$   & \textbf{2.44}$\pm 1.7$   
            &  82.69$\pm 3.0$   &  78.43$\pm 2.6$  &  90.12$\pm 1.4$   & \textbf{2.54}$\pm 1.4$   & \textbf{1.63}$\pm 1.2$   
            & \textbf{78.28}$\pm 1.0$     &  \textbf{86.67}$\pm 1.0$   & 72.00$\pm 1.6$    &  \textbf{5.08}$\pm 3.6$  & \textbf{3.92}$\pm 2.9$  
            \\ \hline
\end{tabular}}
\label{performance}
\end{table*}

\textbf{Effectiveness} The results shown in Table \ref{performance} demonstrate the impressive debiasing power of MAPPING in node classification tasks. 
Compared to vanilla GNNs, $\Delta SP$ and $\Delta EO$ in Table \ref{performance} all decrease, especially in German, GNNs drop more biases than in Recidivism and Credit, wherein GCN introduces more biases than other GNN variants but reduces more biases as well in most cases. As graph size grows, MAPPING still remains promising results and excels all the rest baselines, indicating high scalability.

\textbf{Efficiency} Once pre-debiasing is completed, the rest training time purely reflects the efficiency of vanilla GNNs. Generally, the summed running time of pre-debiasing and GNN training is lower than in-processing methods which introduce extra computational costs, e.g., complex objective functions and/or iterative operations. Even in the small-scale German dataset, the whole running time of MAPPING is always less than $150$ seconds for $10$ trials, but FairGNN \cite{dai2021sayno_fair} and NIFTY \cite{nifty} (excluding counterfactual fairness computation) are $1.17$-$9.28$ times slower. Since we directly use EDITS's \cite{dong2022edits} debiased datasets, there is no comparison for pre-debiasing, but EDITS is $1.00$-$6.43$ times slower in running GNNs. We argue that MAPPING modifies more features and edges but still achieves competitive debiasing and classification performance. Concerning time complexity, the key lies in $dCov$ and $dCor$ calculations, which is $\mathcal{O}(|\mathcal{V}|^{2})$ \cite{dcor_complexity}. For feature debiasing, the time complexity of pre-masking is $\mathcal{O}(d|\mathcal{V}|^{2})$, where $d$ is the dimension of original features. Since $d$ is small, the time complexity is comparable to $\mathcal{O}(|\mathcal{V}|^{2})$; the time complexity of reconstruction for each training epoch is $2\mathcal{O}(|\mathcal{V}|^{2})+\mathcal{O}(d)$, which is comparable to $\mathcal{O}(|\mathcal{V}|^{2})$. As for topology debiasing, fair MP is in time complexity of $\mathcal{O}(|\mathcal{V}|^{2})+\mathcal{O}(|\mathcal{V}|)$ for each training epoch and post-pruning is $\mathcal{O}(|\mathcal{E}|)$. As suggested in \cite{fast_dcov}, the time complexity of $dCov$ can be further reduced to $\mathcal{O}(|\mathcal{V}|\log(|\mathcal{V}|))$ for univariate cases. 

\subsubsection{\textbf{Framework Flexibility and Model Performance}}

To answer \textbf{Q2} to \textbf{Q4}, we compare MAPPING against other baselines and launch attribute inference attacks to investigate the effects of sensitive information leakage mitigation of MAPPING.

\textbf{Flexibility} To answer \textbf{Q2}, from Table \ref{performance}, we observe that compared to vanilla GNNs, MAPPING improves utility in most cases, especially training German with GIN, and Recidivism with GCN and GraphSAGE. We argue that MAPPING can remove features that contribute less to node classification and meanwhile delete redundant and noisy edges. Plus the debiasing analysis, we conclude that MAPPING can flexibly adapt to diverse types of GNN variants. 

\textbf{Model Comparison and Trade-offs} To answer \textbf{Q3}, we observe that MAPPING achieves more competitive performance than other SOTA models. In view of utility, MAPPING more or less outperforms in one or more utility metrics in all cases. It even outperforms all other baselines in all metrics when training German with GIN and Recidivism with GCN. With respect to fairness, on most occasions, all debiasing models can effectively alleviate biases, wherein MAPPPING outperforms others. Moreover, MAPPING is more stable than the rest. 
Overall, we conclude that MAPPING achieves better utility and fairness trade-offs than the baselines. 

\begin{figure}[htbp]
    \centering
    \subfloat[]{
    \includegraphics[width=0.45\linewidth]{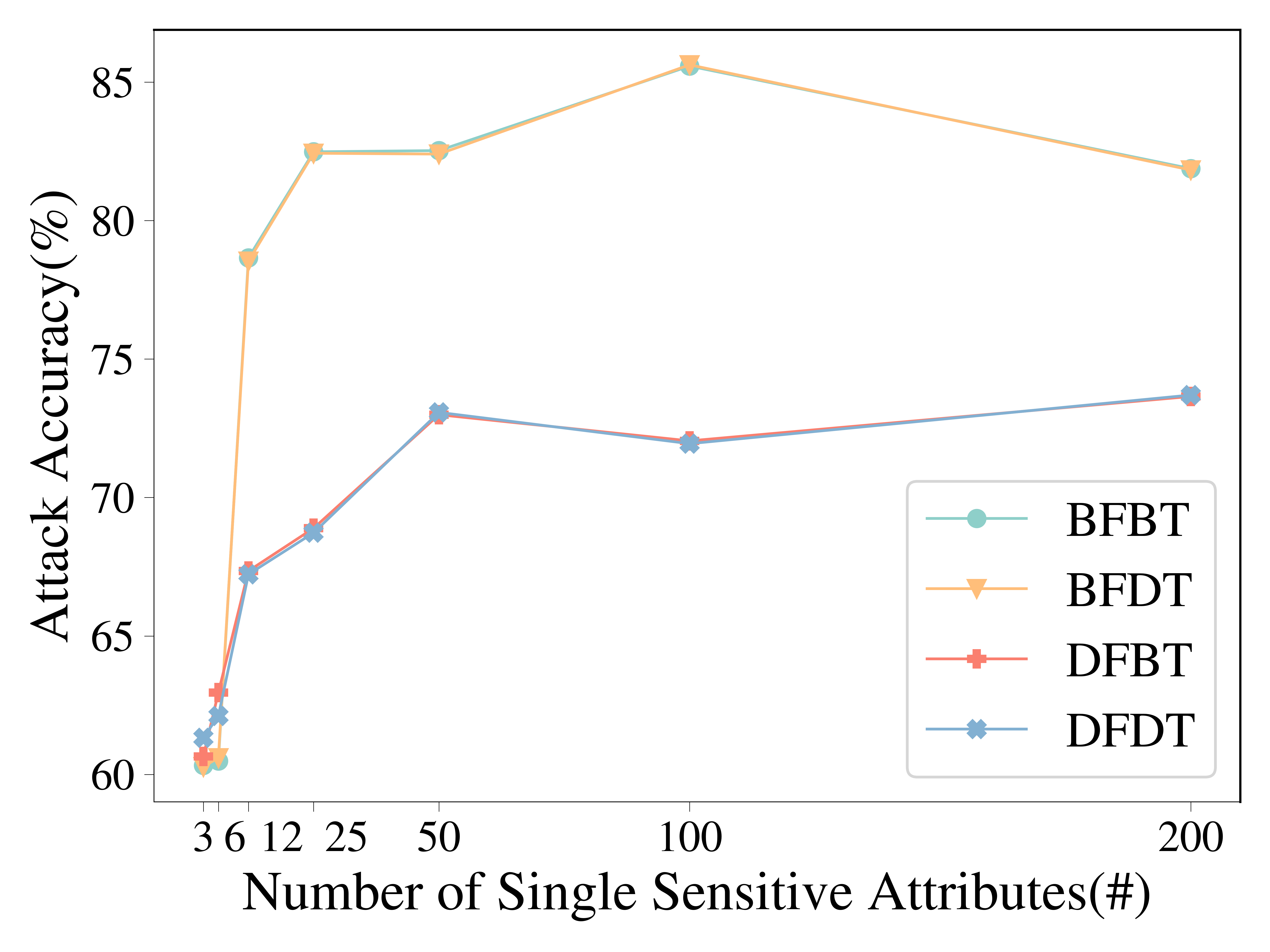}
    \label{acc_german}} 
    \subfloat[]{
    \includegraphics[width=0.45\linewidth]{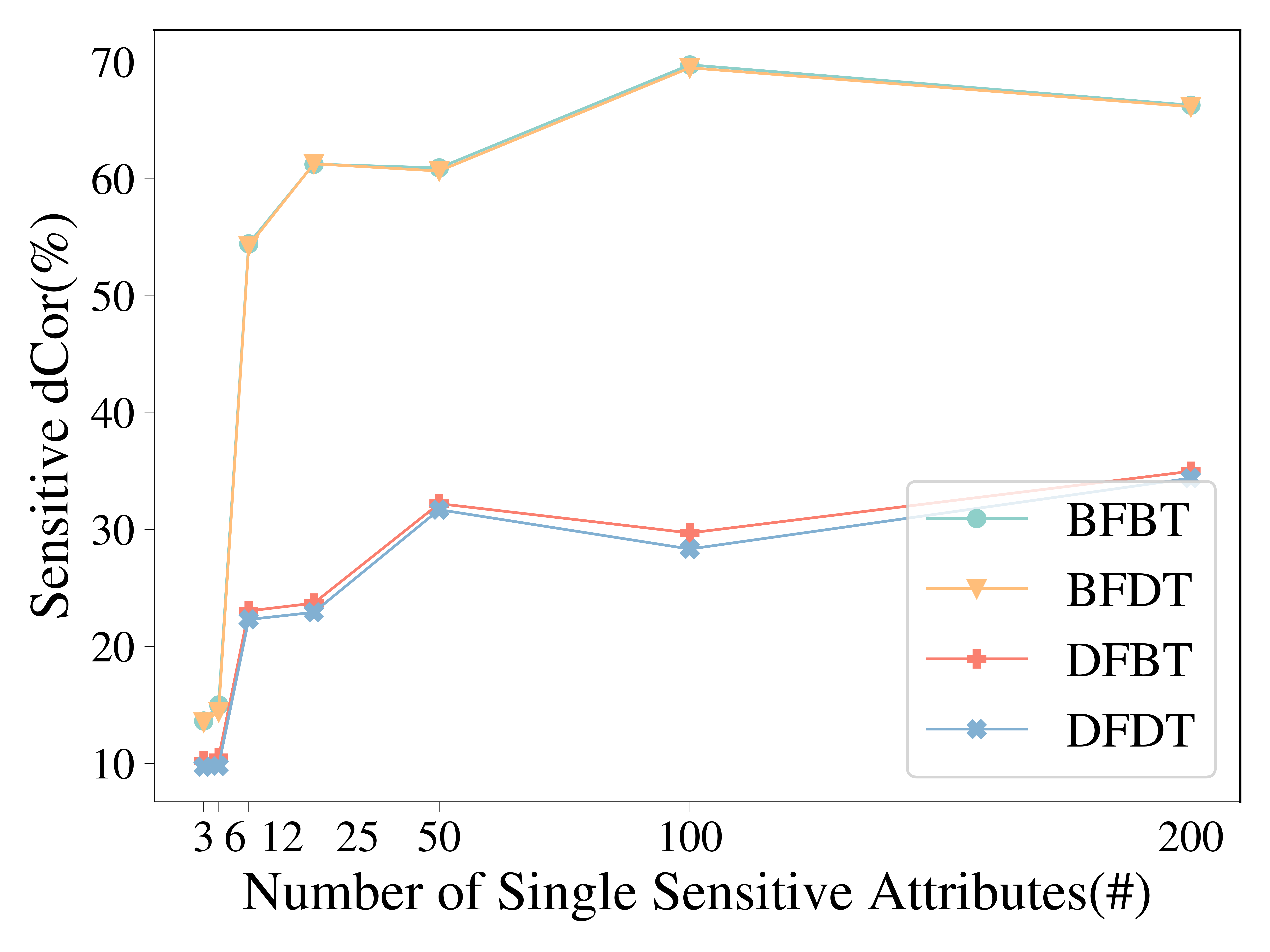}
    \label{sens_dcor_german}}
       \caption{Attribute Inference Attack Under Different Inputs}
    \label{fig:att_german}
\end{figure}

\textbf{Sensitive Information Leakage} To answer \textbf{Q4}, since German reduces the largest biases, we simply use German to explore sensitive information leakage under different input pairs. As shown in Figure \ref{fig:att_german}, since all sensitive attributes are masked and MAPPING only prunes a small portion of edges in German, the attack accuracy and sensitive correlation of BFBT pair are quite close to the BFDT pair while DFDT's performance is slightly lower than the DFBT pair, which further verifies the aforementioned empirical findings and elucidates that MAPPING can effectively confine attribute inference attacks even when adversaries can collect large numbers of sensitive labels. Please note that there are some performance drops, we argue it is due to the combined effects of data imbalance and more stable performance after collecting more sensitive labels. 
\subsection{\textbf{Extension to Multiple Sensitive Attributes}} We set the main sensitive attribute as gender and the minor as age $\left(\le 25/>25\right)$. Still, we adopt German to perform evaluation and explore sensitive information leakage. The details follow the same pattern in Subsection \ref{perf}. Since the SOTA cannot be trivially extended to multiple sensitive attribute cases, we only compare the performances of the vanilla and MAPPING. Besides, after careful checking, we found that only `gender' and `age' can be treated as sensitive attributes, the other candidate is `foreigner', but this feature is too vague, and cannot indicate the exact nationality, so we solely use the above two sensitive attributes for experiments. And the hyperparameters are exactly the same in the experimental section.
\begin{table*}[!htbp]
\caption{\textbf{Node Classification of Multiple Sensitive Attributes}}
\resizebox{\linewidth}{!}{
\begin{tabular}{clccccccc}
\hline

                          \textbf{GNN}  & \textbf{Variants} & \textbf{ACC}  & \textbf{F1}  & \textbf{AUC} & \textbf{\bm{$\Delta SP_{minor}$}} & \textbf{\bm{$\Delta EO_{minor}$}} & \textbf{\bm{$\Delta SP_{major}$}} & \textbf{\bm{$\Delta EO_{major}$}} \\ \hline
\multirow{2}{*}{\textbf{GCN}} 
            & Vanilla                    
            & 65.68$\pm 8.7$     & 70.24$\pm 12.0$    & \textbf{74.39}$\pm 0.5$    &  28.60$\pm 5.2$  & 28.65$\pm 3.9$  &
            36.19$\pm 5.0$  & 28.57$\pm 1.6$\\

            & MAPPING                
            & \textbf{69.52}$\pm 5.7$     & \textbf{78.82}$\pm 9.1$    & 73.23$\pm 0.9$    &  \textbf{4.92}$\pm 5.1$  & \textbf{5.05}$\pm 7.0$  &
            \textbf{9.30}$\pm 5.0$  & \textbf{6.46}$\pm 4.3$\\  \hline
            
\multirow{2}{*}{\textbf{GraphSAGE}} 
            & Vanilla
            & \textbf{71.64}$\pm 1.1$     & \textbf{81.46}$\pm 1.1$    & \textbf{71.07}$\pm 2.4$    &  15.88$\pm 9.0$  & 10.47$\pm 8.0$  &
            19.47$\pm 9.1$  & 11.78$\pm 7.0$ \\
            
            & MAPPING                
            & 67.40$\pm 10.4$     & 74.28$\pm 19.2$    & 70.69$\pm 1.6$    &  \textbf{7.07}$\pm 5.0$  & \textbf{6.35}$\pm 4.5$  &
            \textbf{11.67}$\pm 8.5$  & \textbf{6.42}$\pm 4.8$ \\   \hline
            
\multirow{2}{*}{\textbf{GIN}}    
            & Vanilla
            & 70.04$\pm 0.1$     & 81.98$\pm 1.2$    & 68.37$\pm 4.5$    &  2.17$\pm 6.0$  & 2.13$\pm 6.2$  &
            2.99$\pm 8.9$  & 2.14$\pm 6.2$     \\
            
            & MAPPING                
            & \textbf{70.56}$\pm 1.1$     & \textbf{82.24}$\pm 0.5$    & \textbf{72.17}$\pm 1.7$    &  \textbf{0.55}$\pm 1.3$  & \textbf{1.07}$\pm 2.7$  &
            \textbf{1.90}$\pm 4.2$  & \textbf{1.12}$\pm 2.2$  \\ \hline
\end{tabular}}
\label{multi_eva}
\end{table*}

From Table \ref{multi_eva}, we can observe that MAPPING achieves powerful debiasing effects even when training with GIN which the original model contains very small biases. MAPPING likewise owns better utility performance when training with GCN and GIN. Whether debiasing or not, GraphSAGE is less consistent and stable, but MAPPING does debias almost up to $50\%$. Besides, the major sensitive attributes are always more biased than the minor. Please note that we hierarchically adopt exactly the same hyperparameters, fine-tuning is not used, hence we cannot promise relatively optimal results. However, this setting has generated sufficiently good trade-offs between utility and fairness.

\label{multi}
\begin{figure}[H]
    \centering
    \subfloat[Attack Accuracy of Multiple Sensitive Attributes]
    {
        \includegraphics[width=0.45\textwidth]{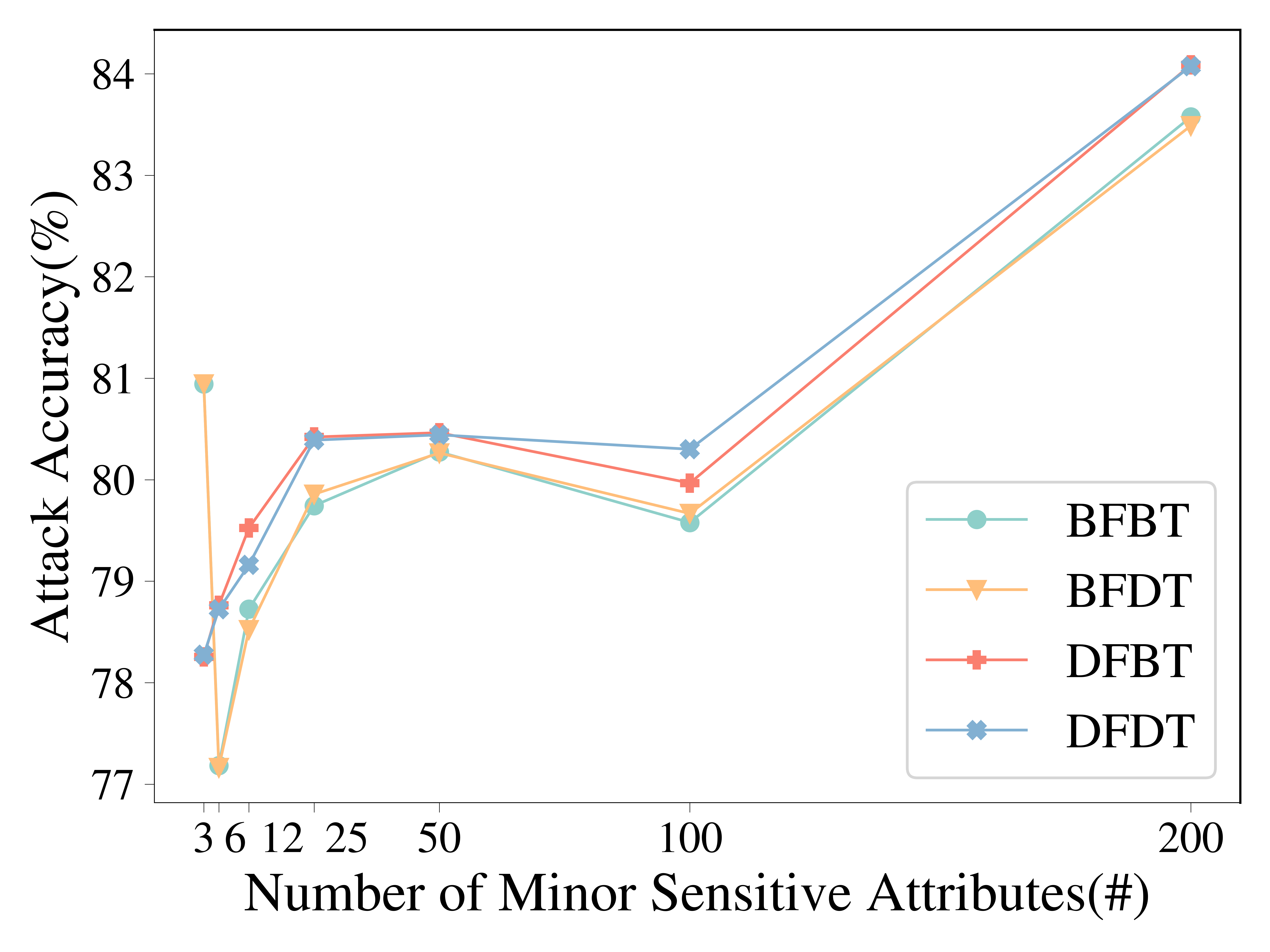} 
        \includegraphics[width=0.45\textwidth]{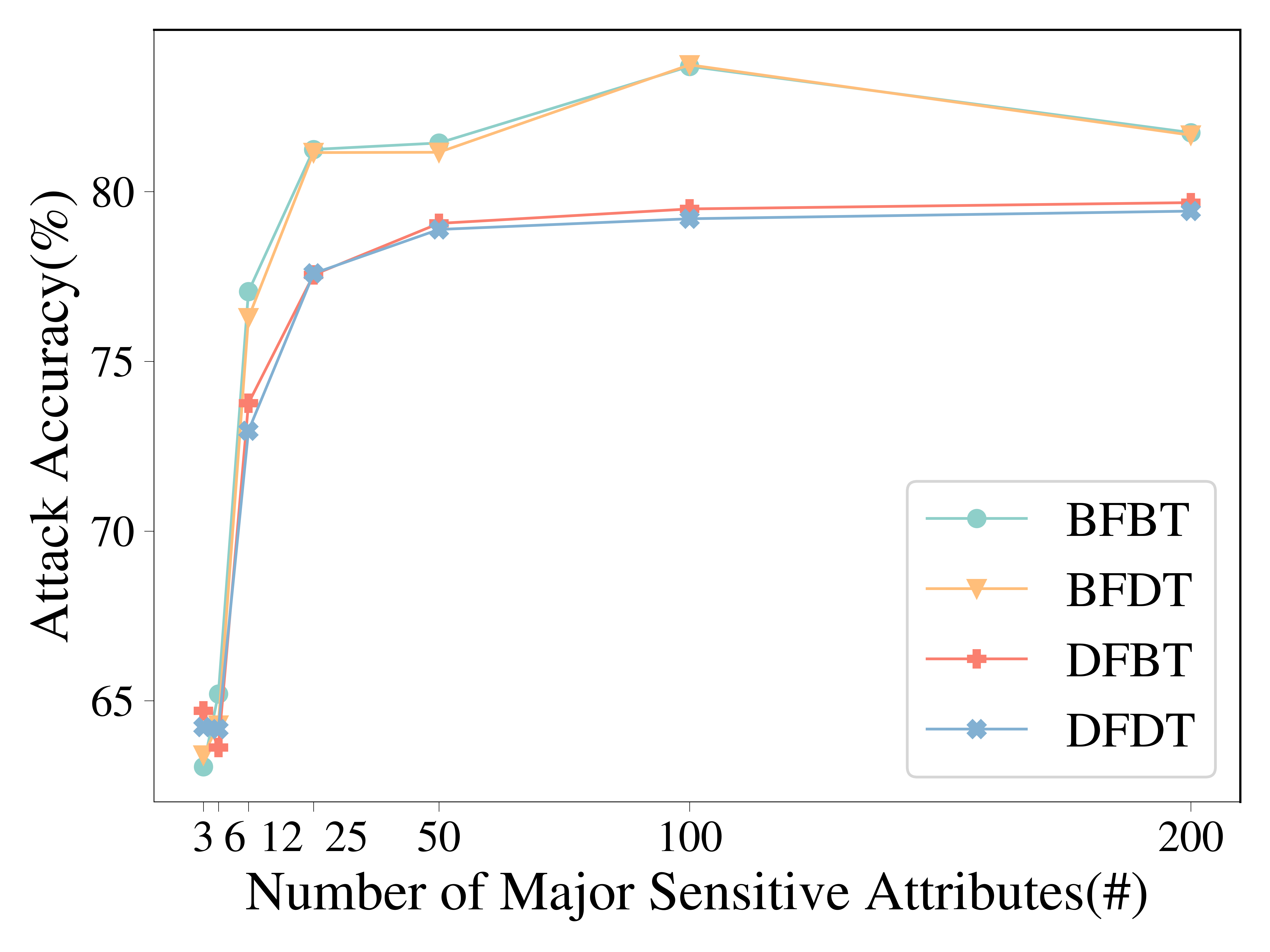}
        \label{attmul}}
    \hfill
    \subfloat[Sensitive Correlation of Multiple Sensitive Attributes]
    {
        \includegraphics[width=0.45\textwidth]{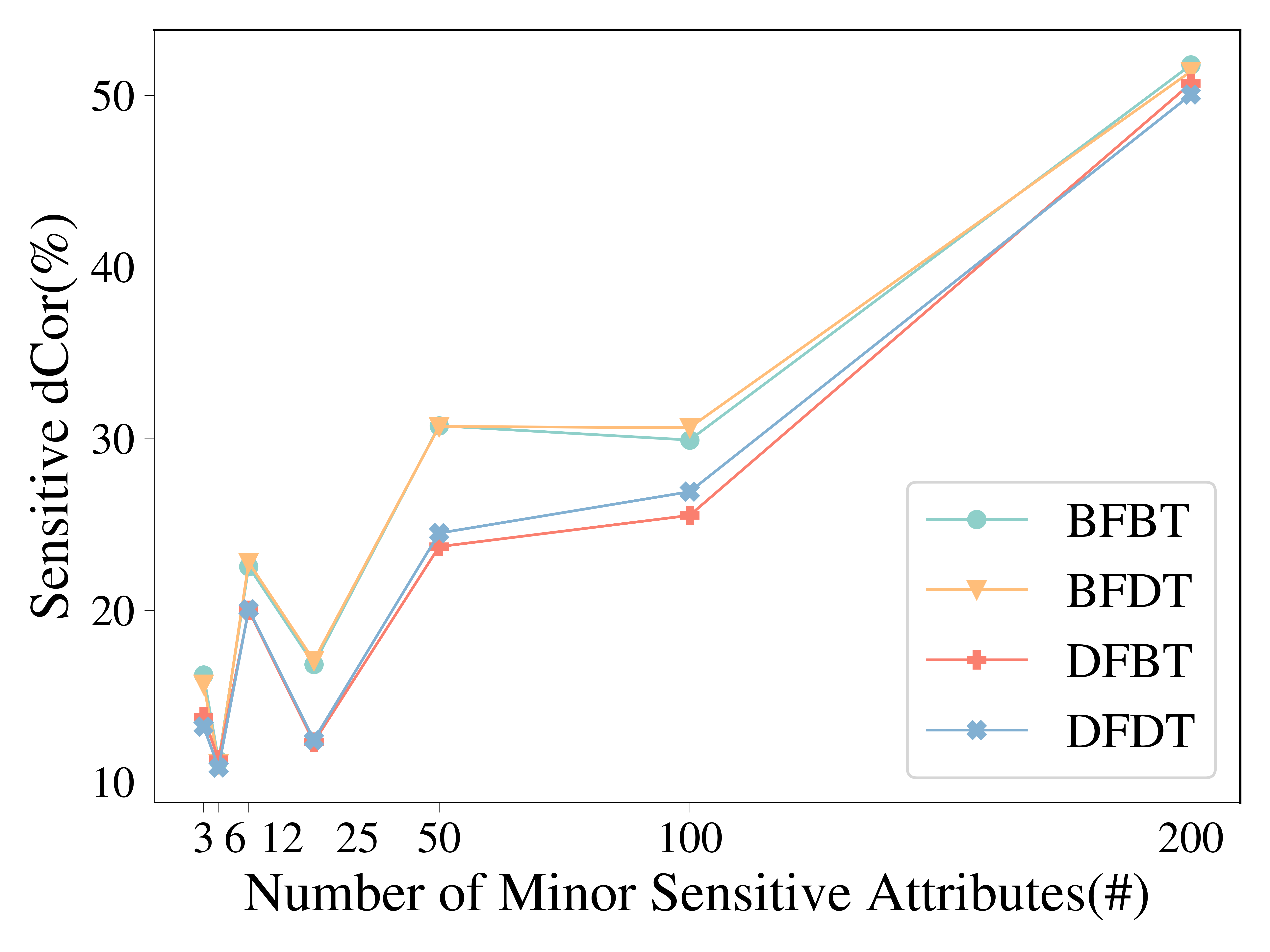} %
        \includegraphics[width=0.45\textwidth]{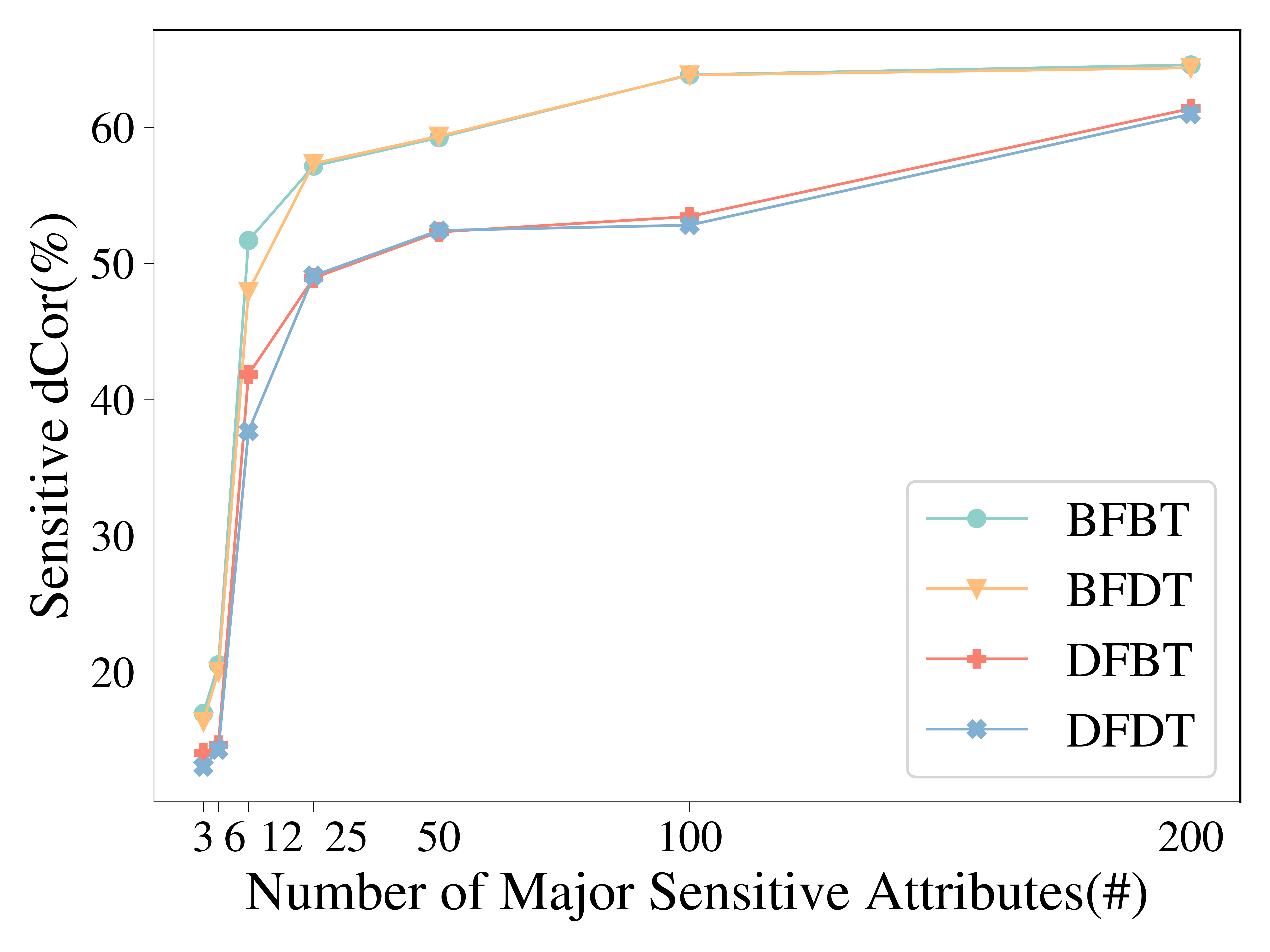}
        \label{attdcor}}
    \caption{Attribute Inference Attack Under Different Inputs in the Multiple Case}
    \label{multiple}
\end{figure}

From Figure \ref{multiple}, we observe similar patterns in the previous experiments. E.g, the performances of BFBT and BFDT pairs are close while the DFBT and DFDT pairs are close; and the unstable performances in fewer label cases may be the main reason why there are some turning points. The more interesting phenomenon is the performances of all pairs of the minor sensitive attributes are pretty close to each other, and their attack accuracy and sensitive correlation increase rapidly after collecting more sensitive labels, but the performances of all pairs of the major increase dramatically at the fewer label situations and seem stabilized to some points. Its sensitive correlation is higher than the minor but the attack accuracy is much lower. We leave the deeper investigation of this inverse fact as the future work.
\subsection{Impact Studies} To answer \textbf{Q5}, we conduct ablation and parameter studies to explore how each debiasing process in MAPPING contributes to fair node classification. As before, we solely adopt German to illustrate the impact of each process.

\subsubsection{\textbf{Ablation Studies}}
MAPPING is composed of two modules and corresponding four processes, namely, pre-masking, and reconstruction for the feature debiasing module, and fair MP and post-pruning for the topology debiasing module. 
We test different MAPPING variants from down-top perspectives, i.e, we first train without pre-masking (w/o-msk), then train without reconstruction (w/o-re), and finally train without the feature debiasing module (w/o-fe). However, the pipeline is different for topology debiasing. Since post-pruning is tightly associated with fair MP, updated edge weights may not be all equal to $0$, which requires considering all possible edges. To avoid undesired computational costs, we treat training without post-pruning and training without topology debiasing module (w/o-to) as the same case. In turn, if we directly remove fair MP, there is no edge to modify. Thus we purely consider training without the topology debiasing module.

As shown in Table \ref{ablation}, w/o-msk or w/o-re leads to more biases and w/o-fe results in the most biases, while w/o-to introduces less bias than other variants. In most cases, different variants sacrifice fairness for utility. The results verify the necessity of each module and corresponding processes to alleviate biases, and demonstrate comparable performance in utility. 
\begin{table*}[!htbp]
\caption{\textbf{Ablation Studies on German}}
\resizebox{\linewidth}{!}{
\begin{tabular}{clccccc}
\hline

                          \textbf{GNN}  & \textbf{Variants} & \textbf{ACC}  & \textbf{F1}  & \textbf{AUC} & \textbf{\bm{$\Delta$}SP} & \textbf{\bm{$\Delta$}EO} \\ \hline
\multirow{6}{*}{\textbf{GCN}} 
            & Vanilla                    
            & 72.00$\pm 2.8$     & 80.27$\pm 2.5$    & \textbf{74.34}$\pm 2.4$    &  31.42$\pm 9.5$  & 22.56$\pm 6.2$ \\
            
            & w/o-msk                      
            &  71.56$\pm 0.7$    &  80.93$\pm 1.0$   & 72.60$\pm 1.6$    &  17.37$\pm 7.7$ &  13.26$\pm 6.6$ \\
                           


            & w/o-re
            &  68.56$\pm 12.4$    &  73.37$\pm 22.7$   &  72.74$\pm 3.5$   &  16.82$\pm 6.5$  & 10.67$\pm 4.8$   \\
            
            & w/o-fe                   
            &  \textbf{72.40}$\pm 2.0$     &  \textbf{81.37}$\pm 1.3$   & 72.86$\pm 4.3$    & 25.70$\pm 17.0$   &  18.07$\pm 11.8$  \\
                           

            & w/o-to                
            & 70.44$\pm 1.6$     & 80.03$\pm 1.8$    & 72.05$\pm 1.0$    &  5.90$\pm 2.3$  & 4.20$\pm 2.3$ \\ 

            & MAPPING                
            & 70.84$\pm 1.8$     & 81.33$\pm 1.3$    & 70.86$\pm 1.9$    &  \textbf{4.54}$\pm 2.2$  & \textbf{4.00}$\pm 1.7$ \\  \hline
            
\multirow{6}{*}{\textbf{GraphSAGE}} 
            & Vanilla                    
            & 71.76$\pm 1.4$     & 81.86$\pm 0.8$    & 71.10$\pm 3.1$    & 14.00$\pm 8.4$   & 7.10$\pm 4.9$  \\

            & w/o-msk                      
            &  70.52$\pm 1.9$    &  81.19$\pm 0.9$   & 69.46$\pm 3.9$    &  8.79$\pm 7.7$  &  5.16$\pm 5.0$ \\
                           


            & w/o-re
            &  72.36$\pm 2.0$    &  82.34$\pm 1.1$   &  71.08$\pm 2.2$   &  8.88$\pm 6.2$  & 4.20$\pm 2.6$   \\
            
            & w/o-fe                   
            &  \textbf{72.84}$\pm 1.1$     &  \textbf{82.39}$\pm 0.7$   & \textbf{71.87}$\pm 3.0$    & 16.75$\pm 7.8$   &  9.59$\pm 5.4$  \\
                           

            & w/o-to                
            & 70.76$\pm 1.2$     & 81.50$\pm 0.6$    & 68.75$\pm 4.1$    &  4.94$\pm 4.8$  & 2.80$\pm 2.7$ \\ 
            
            & MAPPING                
            & 70.76$\pm 1.2$     & 81.51$\pm 0.7$    & 69.89$\pm 1.9$    & \textbf{3.73}$\pm 3.1$   & \textbf{2.39}$\pm 1.5$ \\   \hline
            
\multirow{6}{*}{\textbf{GIN}}       
            & Vanilla                    
            & 71.88$\pm 1.5$     & 81.93$\pm 0.7$    & 67.21$\pm 10.3$    &  14.07$\pm 10.6$  & 9.78$\pm 8.2$     \\
            
            & w/o-msk                      
            &  73.24$\pm 1.1$    &  83.25$\pm 0.5$   & 71.49$\pm 2.0$    &  7.22$\pm 4.7$  &  2.68$\pm 4.1$ \\
                           


            & w/o-re
            &  73.96$\pm 2.1$    &  83.46$\pm 0.8$   &  70.79$\pm 4.4$   &  3.46$\pm 2.5$  & \textbf{1.68}$\pm 1.6$   \\
            
            & w/o-fe                   
            &  \textbf{74.08}$\pm 0.8$     &  \textbf{83.53}$\pm 0.2$   & \textbf{73.02}$\pm 0.6$    & 18.10$\pm 8.1$   &  9.82$\pm 7.2$  \\
                           

            & w/o-to                
            & 72.92$\pm 1.9$     & 82.88$\pm 1.0$    & 70.49$\pm 1.1$    &  \textbf{2.19}$\pm 1.5$  & 3.21$\pm 2.2$ \\ 
            
            & MAPPING                
            & 73.40$\pm 1.2$     & 83.18$\pm 0.5$    & 71.48$\pm 0.6$    & 2.20$\pm 1.4$   & 2.44$\pm 1.7$  \\ \hline
\end{tabular}}
\label{ablation}
\end{table*}

\subsubsection{\textbf{Parameter Studies}} 
We mainly focus on the impacts of fairness-relevant coefficients $\lambda_2$, $\lambda_3$ for feature debiasing and $\lambda_4$ and $r_p$ for topology debiasing. The original choices are 3.50e4, 0.02, 1.29e4, 0.65 for German, 5e4, 100, 515 and 0.72 for Recidivism, and 8e4, 100, 1.34e5 and 0.724 for Credit, respectively. Still, we employ German to illustrate. Please note that we fix the pruning threshold $r_p$ and only investigate coefficients in objective functions. We vary $\lambda_2 \in$ \{0, 1e-5, 1e-3, 1, 1e3, 1e5, 1e7\} when $\lambda_3$ and $\lambda_4$ are fixed, and alter $\lambda_3 \in$ \{0, 1e-7, 1e-5, 1e-3, 1, 1e3, 1e5\} when $\lambda_2$ and $\lambda_4$ are fixed, and finally change $\lambda_4 \in$\{0, 1e-5, 1e-3, 1, 1e3, 1e5, 1e7\} when $\lambda_2$ and $\lambda_3$ are fixed. Please note different colors represent for different coefficients, i.e., the green line denotes $\lambda_2$ and the red and blue lines denote $\lambda_3$ and $\lambda_4$, respectively. Different types of lines represent for different GNN variants, i.e., the straight line denotes GCN and the dotted line with larger intervals and the dotted line with smaller intervals denote GraphSAGE and GIN, respectively.
\label{paras}
\begin{figure}[H]
    \centering
    \subfloat[Utility Performance of Node Classification]
    {
        \includegraphics[width=0.45\textwidth]{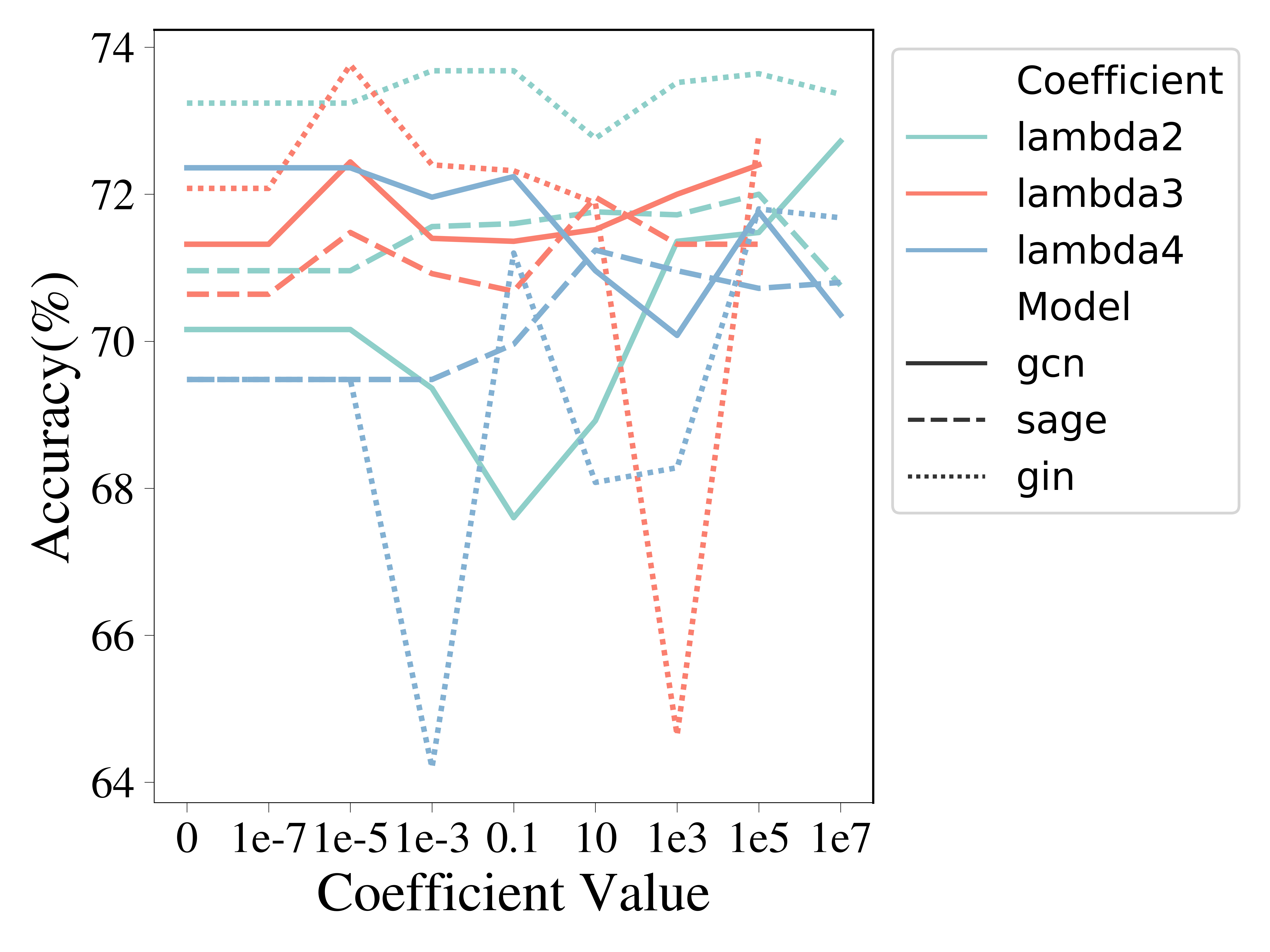} 
        \includegraphics[width=0.45\textwidth]{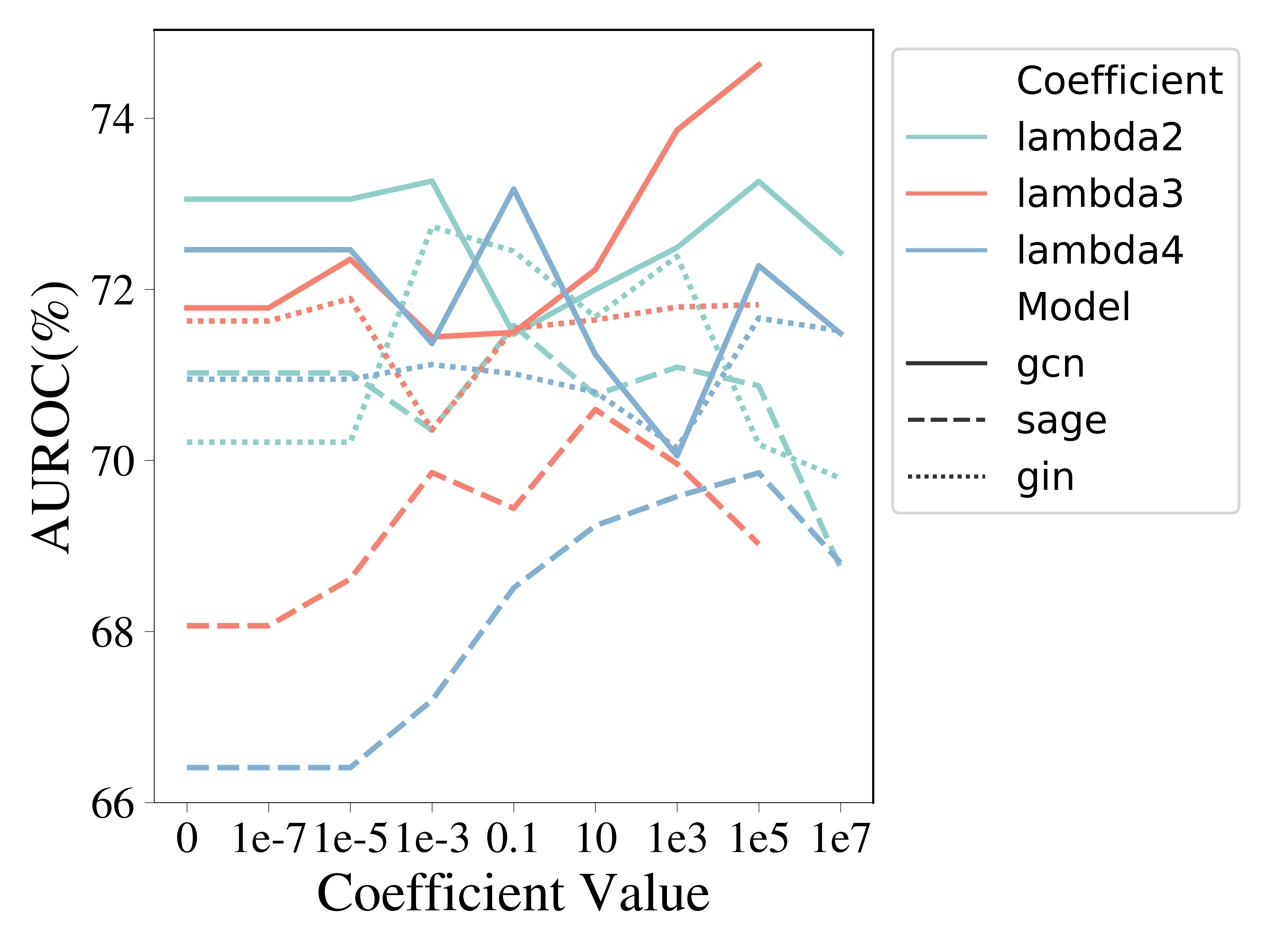}
        \label{subfig:impact_utility}}
    \hfill
    \subfloat[Fairness Performance of Node Classification]
    {
        \includegraphics[width=0.45\textwidth]{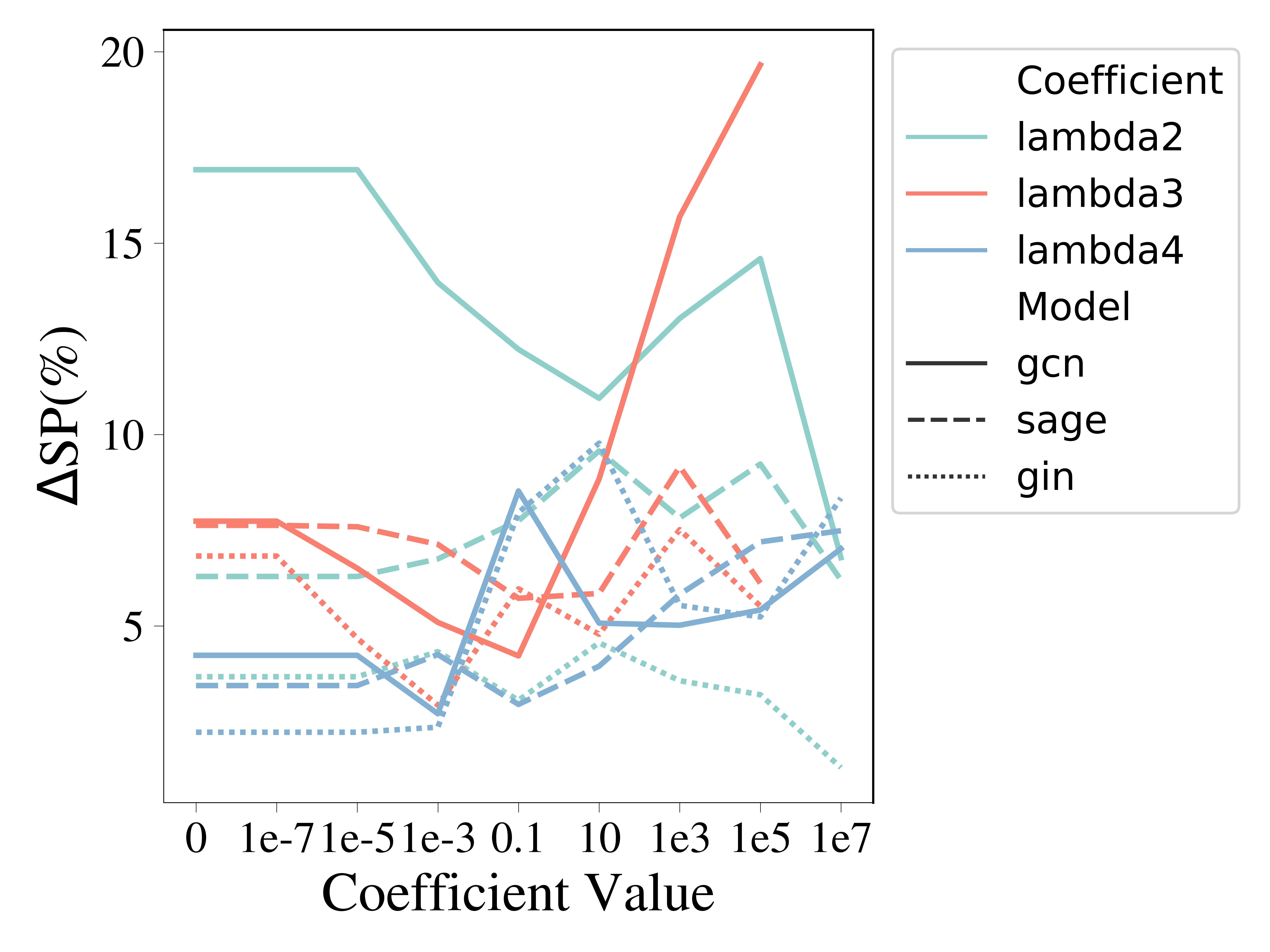} %
        \includegraphics[width=0.45\textwidth]{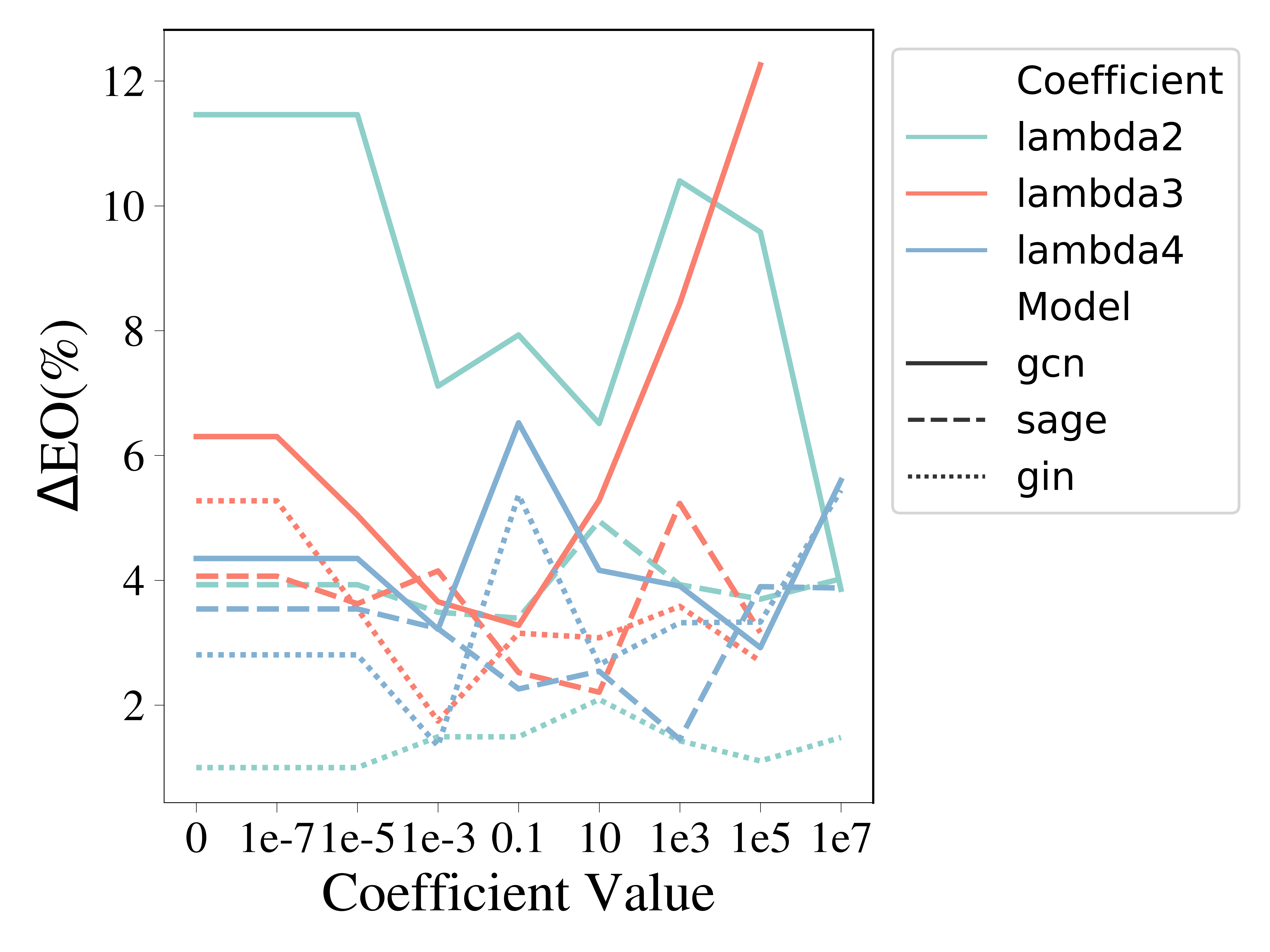}
        \label{subfig:impact_fair}}
    \caption{Coefficient Impact of Utility and Fairness}
    \label{impact_mapping}
\end{figure}
As shown in Figure \ref{impact_mapping}, the different choices from wide ranges all mitigate biases, and sometimes they can reach win-win situations for utility and fairness, e.g, $\lambda_2$=1e5. Besides, they can achieve better trade-offs between utility and fairness when $\lambda_2\in$ [1e3,1e5], $\lambda_3\in$ [1e-3,10] and $\lambda_4\in$ [1e3,1e5] for all GNN variants.

\subsection{Limitations and Future Work}
Although MAPPING demonstrates promising debiasing effects and utility performance with confined multiple sensitive information leakage, we acknowledge several limitations that can shed light on future work.

We solely deploy debiasing models without consideration of privacy-enhancing techniques, as our goal is to investigate how fairness intervention interacts with multiple attribute privacy. A potential future direction is to develop a unified, model-agnostic debiasing framework with rigorous privacy guarantees, such as differential privacy, and explore the fairness impacts of privacy-protection techniques to foster the development of trustworthy GNNs.

As this work primarily focuses on empirical analysis, another potential direction is to provide theoretical support to uncover some interesting patterns under multiple sensitive attribute cases. Additionally, determining theoretically optimal values for sensitive thresholds and coefficients in objective functions, rather than relying on experience or empirical tests, would be helpful.

It would be valuable to gather existing literature and categorize all possible fairness constraints into distinct types, and then incorporate each constraint into MAPPING, evaluate its performance and provide deeper analysis on why distance correlation/covariance-based constraint outperform others, especially in multiple sensitive attribute cases. 

MAPPING demonstrates its effectiveness and efficiency across graphs of varying scales, though very large graphs are beyond the scope of this study. However, given the size of real-world graphs, e.g., social networks, it is essential to explore how MAPPING can be applied to very large graphs. One potential direction is to employ some pre-processing techniques such as graph sampling and graph reduction, and then leverage MAPPING as usual.

%% file: Related.tex
\section{Related Work}
In this section, we summarize representative work close to MAPPING. We refer the interested readers to \cite{dong_fairgraph_survey, li2022private} for extensive surveys on fair and private graph learning.

\textbf{Fair or Private Graph Learning}
For fair graph learning, at the pre-processing stage, EDITS \cite{dong2022edits} is the first work to construct a model-agnostic debiasing framework based on $Was$, which reduces feature and structural biases by feature retuning and edge clipping. At the in-processing stage, FairGNN \cite{dai2021sayno_fair} first combines adversarial debiasing with $Cov$ constraints to learn a fair GNN classifier under missing sensitive attributes. NIFTY \cite{nifty} augments graphs from node, edge and sensitive attribute perturbation respectively, and then optimizes by maximizing the similarity between the augmented graph and the original one to promise counterfactual fairness in node representation. At the post-processing stage, FLIP \cite{burst_filter_bubble} achieves fairness by reducing graph modularity with a greedy algorithm, which takes the predicted links as inputs and calculates the changes in modularity after link flipping. But this method is only adapted to link prediction tasks. Excluding strict privacy protocols, existing privacy-preserving studies only partially aligns with fairness, e.g., \cite{Adversarial_privacy_preserving_against_infer, info_obfuscation_privacy_protection} employ attackers to launch attribute inference attacks and utilize game theory to decorrelate biases from node representations. Another line of research \cite{privacy_protection_partial_sens_attr,info_obfuscation_privacy_protection,GNN_mutual_privacy} introduce privacy constraints such as orthogonal subspace, $Was$ or $MI$ to remove linear or mutual dependence between sensitive attributes and node representations to fight against attribute inference or link-stealing attacks.

\textbf{Interplays between Fairness and Privacy on Graphs}
Since few prior work address interactions between fairness and privacy on graphs, i.e., rivalries, friends, or both, we first enumerate representative research on i.i.d. data. 
One research direction is to explore privacy risks and protection of fair models. Chang et.al. \cite{privacy_risk_fair} empirically verifies that fairness gets promoted at the cost of privacy, and more biased data results in the higher membership inference attack risks of achieving group fairness; FAIRSP \cite{fairnessmeetprivacy} shows stronger privacy protection without debiasing models leads to better fairness performance while stronger privacy protection in debiasing models will worsen fairness performance. Another research direction is to investigate fairness effects under privacy guarantee, e.g., differential privacy \cite{dp_original}, which typically exacerbate disparities among different demographic groups \cite{fairwithprivacyprotection,empiricalfairprivacy} without fairness interventions. While some existing studies \cite{dpfair, sayno_privacy_extension} propose unified frameworks to simultaneously enforce fairness and privacy, they do not probe into detailed interactions. 
PPFR \cite{interaction_priv_fair} is the first work to empirically show that the privacy risks of link-stealing attacks can increase as individual fairness of each node is enhanced. Moreover, it models such interplays via influence functions and $Cor$, and finally devises a post-processing retraining method to reinforce fairness while mitigating edge privacy leakage. To the best of our knowledge, there is no thorough prior GNN research to address the interactions at the pre-processing stage. 

%% file: Conclusion.tex
\section{Conclusion}
In this research, we take the first major step toward exploring the inner relationship between group fairness and multiple attribute privacy in GNNs at the pre-processing stage. We empirically demonstrate that GNNs not only preserve and amplify biases but also exacerbate the leakage of multiple sensitive attributes. This observation motivates us to propose a novel model-agnostic debiasing framework named MAPPING. Specifically, MAPPING utilizes $dCov$-based fairness constraints and adversarial training to jointly debias both features and topologies while mitigating inference risks of multiple sensitive attributes. Our empirical experiments confirm the effectiveness and flexibility of MAPPING, as it achieves superior trade-offs between utility and fairness, while simultaneously limiting sensitive information leakage, thereby contributing to the development of trustworthy GNNs.

%% file: Appendix.tex
\section{Empirical Analysis}

\subsection{Data Synthesis}
\label{synthesis}

Please note that $S_m$ can be highly related to $S_n$ or even not. For $S_n$, we generate a $2500 \times 3$ biased non-sensitive feature matrix from multivariate normal distributions $\mathcal{N} \left(\mu_0, \Sigma_0\right)$ and $\mathcal{N} \left(\mu_1, \Sigma_1\right)$, where subgroup $0$ represents minority in reality, $\mu_0=\left(-10,-2,-5\right)^{T}$, $\mu_1=\left(10,2,5\right)^{T}$, $\Sigma_0=\Sigma_1$ are both identity matrices, and $|S_0|=500$ and $|S_1|=2000$. We combine the top $100$ sample from $\left(\mu_0,\Sigma_0\right)$, the top $200$ from $\left(\mu_1,\Sigma_1\right)$, then the next $200$ from $\left(\mu_0,\Sigma_0\right)$ and $600$ from $\left(\mu_1,\Sigma_1\right)$, then the next $100$ from $\left(\mu_0,\Sigma_0\right)$ and $700$ from $\left(\mu_1,\Sigma_1\right)$, and finally the last $100$ from $\left(\mu_0,\Sigma_0\right)$ and $500$ from $\left(\mu_1,\Sigma_1\right)$. Next, generate $S_n$ based on this combination. And then we create another $2500 \times 3$ non-sensitive matrix attached to $S_n$ from multivariate normal distributions $\mathcal{N} \left(\mu_2, \Sigma_2\right)$ and $\mathcal{N} \left(\mu_3, \Sigma_3\right)$, where subgroup $2$ represents minority in reality, $\mu_2=\left(-12,-8,-4\right)^{T}$, $\mu_3=\left(12,8,4\right)^{T}$, $\Sigma_2=\Sigma_3$ are both identity matrices, and $|S_2|=700$ and $|S_3|=1800$. We combine the top $300$ from $\left(\mu_2,\Sigma_2\right)$, and $1200$ from $\left(\mu_3,\Sigma_3\right)$, and then $400$ from $\left(\mu_2,\Sigma_2\right)$ and $600$ from $\left(\mu_3,\Sigma_3\right)$. Next, generate $S_m$ based on this combination again. Debiased features are sampled from multivariate normal distributions with $\mu_4=\left(0,1,0,1,0,1\right)$ and covariance as the identity matrix. Second, the biased topology is formed via the stochastic block model, where the first block contains $500$ nodes while the second contains $2000$ nodes, and the link probability within blocks is 5e-3 and between blocks is 1e-7. The debiased topology is built with a random geometric graph with $0.033$ radius. 

\begin{figure}[htbp]
    \centering
    \subfloat[Biased non-sensitive features and graph topology (Minor)]
    {
        \includegraphics[width=0.45\textwidth]{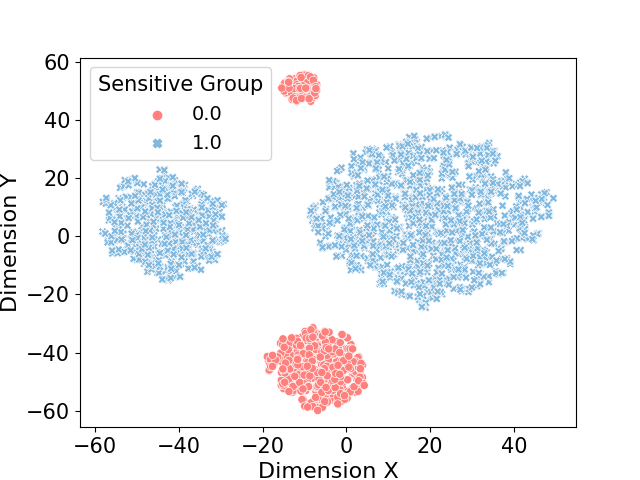}
        \includegraphics[width=0.45\textwidth]{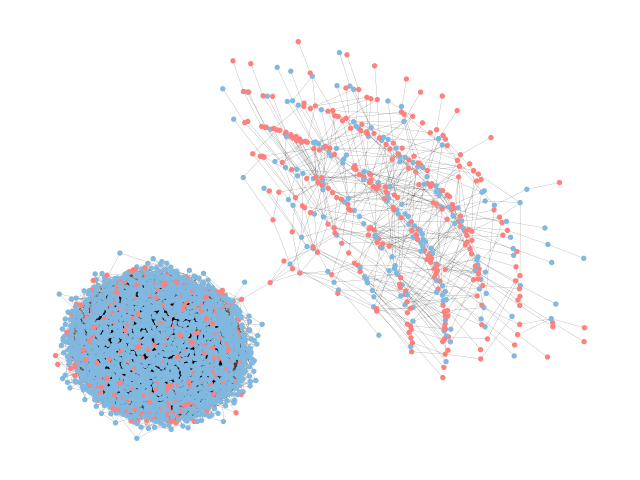}
        \label{Biased features and topology_m}}
    \hfill
    \subfloat[Unbiased non-sensitive features and graph topology (Minor)]
    { 
    \includegraphics[width=0.45\textwidth]{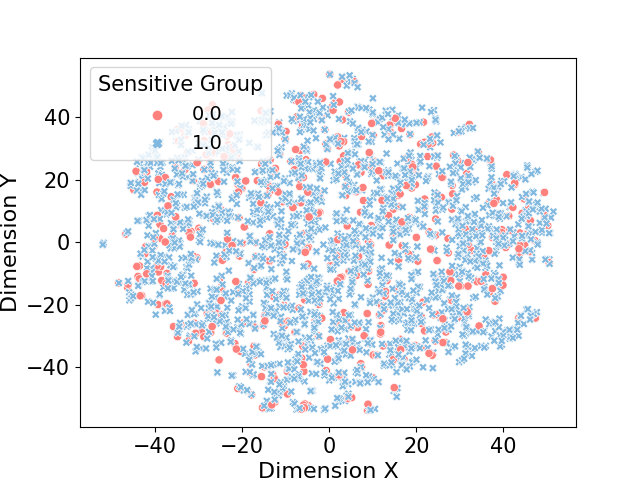}
    \includegraphics[width=0.45\textwidth]{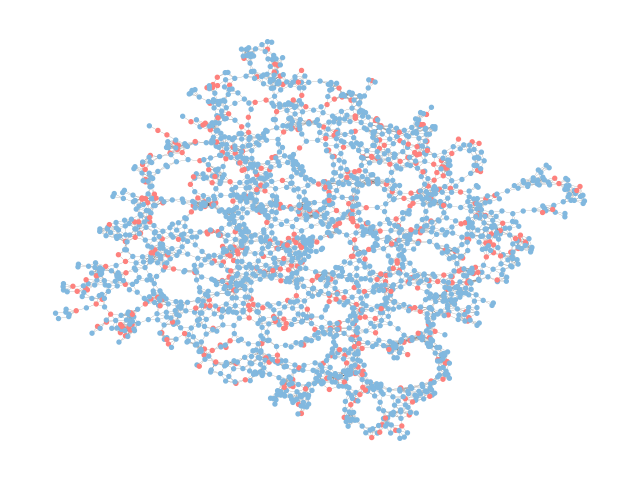}
        \label{Unbiased features and topology_m}}
    \caption{Distributions of Biased and Unbiased Graph Data Based on the Minor Sensitive Attribute. The minor sensitive attribute is binary, where group 0 represents the minority while group 1 denotes the majority.}
    \label{data synthesis}
\end{figure}

\subsection{Implementation Details}
\label{prelim}
We built $1$-layer GCNs with PyTorch Geometric \cite{torch_geometric} with Adam optimizer \cite{adam}, learning rate 1e-3, dropout 0.2, weight decay 1e-5, training epoch 1000, and hidden layer size 16 and implemented them in PyTorch \cite{pytorch}. All experiments are conducted on a 64-bit machine with 4 Nvidia A100 GPUs. The experiments are trained on $1,000$ epochs. We repeat experiments $10$ times with different seeds to report the average results. 

\section{Pseudo Codes}
\label{code}
\subsection{Algorithm 1 - Pre-masking Strategies}
\begin{algorithm}
\caption{Pre-masking Strategies}\label{alg1}
\begin{algorithmic}[1]
\Require Original feature matrix $\mathcal{X}$ with Normal feature matrix $\mathcal{X}_N$ and Sensitive attributes $S$, Ground-truth label $\mathcal{Y}$, distributed ratio $r$, sensitive threshold $r_s$
\Ensure Pre-masked feature matrix $\tilde{\mathcal{X}}$
\State \hspace{0.5cm} Compute $\mathcal{R}_{n}^{2}(\mathcal{X}_i, S)$ and $\mathcal{R}_{n}^{2}(\mathcal{X}_i, \mathcal{Y})$ based on equation(2) \textbf{for} $i\in[1,\dots,d]$;
\State \hspace{0.5cm} Choose the top $x=\lfloor rd\rfloor$ features from descending $\mathcal{R}_{n}^{2}(\mathcal{X}, S)$ to obtain the top related feature set $Set_{top}$; Choose the top $x$ features from ascending $\mathcal{R}_{n}^{2}(\mathcal{X}, \mathcal{Y})$ to obtain the less related feature set $Set_{les}$;
\State \hspace{0.5cm} Obtain the intersection set $Set_{int}=Set_{top}\cap Set_{les}$;
\State \hspace{0.5cm} Filter features whose $\mathcal{R}_{n}^{2}(\mathcal{X}_i, S)<r_s$ to obtain the extremely highly sensitive feature set $Set_{sen}$; 
\State \hspace{0.5cm} Obtain the union set $Set_{uni}=Set_{int}\cup Set_{sen}$;
\State \hspace{0.5cm} Obtain the pre-masked feature matrix $\tilde{\mathcal{X}}=\mathcal{X}\backslash Set_{uni}$.
\State \hspace{0.5cm}\textbf{return} $\tilde{\mathcal{X}}$
\end{algorithmic}
\label{Algo_MAPPING}
\end{algorithm}

\subsection{Algorithm 2 - MAPPING}
\begin{algorithm}
\caption{MAPPING}\label{alg2}
\begin{algorithmic}[1]
\Require Adjacency matrix $A$, Original feature matrix $\mathcal{X}$ with Normal feature matrix $\mathcal{X}_N$ and Sensitive attributes $S$, Ground-truth label $\mathcal{Y}$, MLP for feature debiasing $f_{mlp}$, GNN for topology debiasing $f_{gnn}$
\Ensure Debiased adjacency matrix $\hat{A}$, Debiased feature matrix $\hat{X}$
\State {\textsc{\textbf{Pre-masking}}$(\mathcal{X}):$}
\State \hspace{0.5cm} Implement Algorithm 1; 
\State \hspace{0.5cm}\textbf{return} $\mathcal{\tilde{X}}$
\State
\State {\textsc{\textbf{Reconstruction}}$(\mathcal{\tilde{X}}):$}
\State \hspace{0.5cm} Initialize equal weights $W_{{f0}_{i}}=1$ \textbf{for} $\mathcal{\tilde{X}}_{i}, i\in[1,\dots, d_m]$;
\State \hspace{0.5cm} Train $f_{mlp}(\mathcal{\tilde{X}})$; Update the feature weight matrix $\hat{W}_{f(i)} \gets \hat{W}_{f(i-1)}$ and the debiased feature matrix $\hat{X}(i) \gets \mathcal{\tilde{X}}\hat{W}_{f(i)}$ \textbf{for} $\hat{W}_{f(0)}=W_{f0}$ based on the equation(8) until convergence;
\State \hspace{0.5cm}\textbf{return}  $\hat{X}$
\State
\State {\textsc{\textbf{Fair MP}}$(A,\hat{X}):$}
\State \hspace{0.5cm} Initialize equal weights $W_{{\mathcal{E}_{0}}_{ij}}=1$ \textbf{for} $A_{ij}, i,j\in[1,\dots, n]$;
\State \hspace{0.5cm} Train $f_{gnn}(W_{\mathcal{E}_{0}},A,\hat{X})$; Update the edge weight matrix $\hat{W}_{\mathcal{E}(i)} \gets \hat{W}_{\mathcal{E}(i-1)}$ \textbf{for} $\hat{W}_{\mathcal{E}(0)}=W_{\mathcal{E}_{0}}$ based on the equation(11) until convergence;
\State \hspace{0.5cm}\textbf{return} $\hat{W}_{\mathcal{E}}$
\State
\State {\textsc{\textbf{Post-tuning}}$(\hat{W}_{\mathcal{E}}):$}
\State \hspace{0.5cm} Prune $\hat{W}_{\mathcal{E}}$ based on the equation(12); Obtain $\tilde{W}_{\mathcal{E}}=\hat{A}$;
\State \hspace{0.5cm}\textbf{return} $\hat{A}$
\end{algorithmic}
\label{AlgoMAPPING}
\end{algorithm}

\section{Experiments}
\subsection{Dataset Description}
In German, nodes represent bank clients and edges are connected based on the similarity of clients' credit accounts. To control credit risks, the bank needs to differentiate applicants with good/bad credits. 
In Recidivism, nodes denote defendants who got released on bail at the U.S state courts during 1990-2009 and edges are linked based on the similarity of defendants' basic demographics and past criminal histories. The task is to predict whether the defendants will be bailed. In Credit, nodes are credit card applicants and edges are formed based on the similarity of applicants' spending and payment patterns. The goal is to predict whether the applicants will default. 
\label{data}

\subsection{Hyperparameter Setting of SOTA}
In this subsection, we detail the hyperparameters for different fair models. To obtain relatively better performance, we leverage Optuna to facilitate grid search. 
\label{setting}

\textbf{FairGNN}: dropout from $\{0.1,0.2,0.3,0.4,0.5\}$, weight decay 1e-5, learning rate $\{0.001,0.005,0.01,0.05,0.1\}$, regularization coefficients $\alpha=4$ and $\beta=0.01$, sensitive number $200$ and label number $500$, hidden layer size $\{16,32,64,128\}$, .

\textbf{NIFTY}: project hidden layer size 16, drop edge and feature rates are $0.001$ and $0.1$, dropout $\{0.1,0.3,0.5\}$, weight decay 1e-5, learning rate $\{0.0001,0.001,0.01\}$, regularization coefficient $\{0.4,0.5,0.6,0.7,\\0.8\}$, hidden layer size 16. 

\textbf{EDITS}: we directly use the debiased datasets in \cite{dong2022edits}, dropout $\{0.05,0.1,0.3,0.5\}$, weight decay \{1e-4,1e-5,1e-6,1e-7\}, learning rate $\{0.001,0.005,0.01,0.05\}$, hidden layer size 16.